\documentclass[dvipsnames]{article}
\usepackage{colm2024_conference}

\usepackage{booktabs}
\usepackage{enumitem}
\usepackage{wrapfig}
\usepackage{algorithm}
\usepackage{algpseudocode}
\usepackage{graphicx}
\usepackage[misc]{ifsym}
\usepackage{multicol}
\usepackage{microtype}
\usepackage{amsmath}
\usepackage{colortbl}
\usepackage[utf8]{inputenc}
\usepackage[T1]{fontenc}
\definecolor{lightgray}{rgb}{0.9,0.9,0.9}
\usepackage{caption}
\usepackage{subcaption}
\usepackage{setspace}
\usepackage{url}
\usepackage{multirow}
\usepackage{tabularx}
\usepackage{blindtext}
\usepackage{pgfplots}
\pgfplotsset{compat=1.18}
\usepackage{tikz}
\usetikzlibrary{er,positioning,bayesnet}
\usepackage{makecell}
\usepackage{tipa}
\usepackage{siunitx}
\usepackage{nicefrac}
\usepackage{listings}
\usepackage[raster,skins, most]{tcolorbox}
\usepackage{xltabular}
\usepackage{adjustbox}
\usepackage{xurl}
\usepackage{rotating}
\usepackage{pifont}
\usepackage[normalem]{ulem}

\usepackage{fontawesome}

\usepackage[table]{xcolor}

\useunder{\uline}{\ul}{}

\usepackage{amsmath,amsfonts,bm}

\def\eqref#1{equation~\ref{#1}}
\def\1{\bm{1}}

\DeclareMathAlphabet{\mathsfit}{\encodingdefault}{\sfdefault}{m}{sl}
\SetMathAlphabet{\mathsfit}{bold}{\encodingdefault}{\sfdefault}{bx}{n}

\providecommand{\ghlink}{https://github.com/LOGOS-Hub/LOGOS}

\newcommand*\justify{%
  \fontdimen2\font=0.4em% interword space
  \fontdimen3\font=0.2em% interword stretch
  \fontdimen4\font=0.1em% interword shrink
  \fontdimen7\font=0.1em% extra space
  \hyphenchar\font=`\-% allowing hyphenation
}

\renewcommand{\texttt}[1]{%
  \begingroup
  \ttfamily
  \begingroup\lccode`~=`/\lowercase{\endgroup\def~}{/\discretionary{}{}{}}%
  \begingroup\lccode`~=`[\lowercase{\endgroup\def~}{[\discretionary{}{}{}}%
  \begingroup\lccode`~=`.\lowercase{\endgroup\def~}{.\discretionary{}{}{}}%
  \catcode`/=\active\catcode`[=\active\catcode`.=\active
  \justify\scantokens{#1\noexpand}%
  \endgroup
}

\usepackage{makecell}
\usetikzlibrary{tikzmark}
\makeatletter
\newcommand*\myfontsize{%
  \@setfontsize\myfontsize{7}{8}%
}
\makeatother

\definecolor{uclablue}{RGB}{159, 195, 224}

\definecolor{uclagold}{RGB}{255, 240, 180}

\definecolor{aliceblue}{RGB}{255, 238, 241}

\definecolor{cadmiumgreen}{rgb}{0.0, 0.42, 0.24}

\definecolor{myred}{rgb}{0.7, 0.3, 0.0}
\definecolor{myblue}{rgb}{0.2, 0.3, 0.6}
\definecolor{babygreen}{rgb}{0.85, 0.97, 0.85}

\newcommand\blfootnote[1]{%
  \begingroup
  \renewcommand\thefootnote{}\footnote{#1}%
  \addtocounter{footnote}{-1}%
  \endgroup
}

\definecolor{purple1}{RGB}{126, 107, 196}
\definecolor{purple2}{RGB}{199, 158, 207}
\definecolor{purple3}{RGB}{214, 200, 255}
\definecolor{purple4}{RGB}{254, 240, 255}

\definecolor{deepblue}{RGB}{48, 58, 82}

\definecolor{darkorange}{RGB}{255, 140, 0}
\definecolor{darkblue}{RGB}{84, 112, 198}
\definecolor{lightgreen}{RGB}{145, 204, 117}
\definecolor{lightyellow}{RGB}{250, 200, 88}
\definecolor{lightred}{RGB}{238, 102, 102}
\definecolor{lightblue}{RGB}{115, 192, 222}

\definecolor{darkred}{rgb}{0.55, 0.0, 0.0}
\definecolor{navy}{rgb}{0.0, 0.0, 0.55}
\definecolor{darkgreen}{rgb}{0.0, 0.39, 0.0}

\newtcolorbox{promptbox}{
  breakable,
  colback=blue!10!white,
  arc=5pt,
  boxrule=0.5pt,
  colframe=blue!50!violet,
  before upper={\normalsize},
  fontupper=\fontfamily{ptm}\selectfont,
}

\definecolor{deepPurple}{HTML}{330066}

\definecolor{uclablue_old}{rgb}{0.15, 0.45, 0.68}
\hypersetup{
    breaklinks,
    citecolor=uclablue_old,
    colorlinks=true,
}

\newtcolorbox{mybox}[2][]
  {colback = black!5!white, colframe = black!75!black, fonttitle = \bfseries,
    colbacktitle = black!100!black, enhanced, before upper={\fontsize{8}{11}\obeyspaces\obeylines\selectfont}, fontupper=\selectfont,
    attach boxed title to top left={yshift=-2.2mm,xshift=4mm},
    title=#2,#1}

\title{%
\begin{tabular}[t]{l}
  \parbox[t]{0.8\textwidth}{\centering
    Speaking the Language of Science: \\Toward a General-Purpose Generative Foundation Model for the Natural Sciences
  }
\end{tabular}
}

\author{
Mingyang Li\textsuperscript{1,*}, Yurou Liu\textsuperscript{1,2,*}, Jieping Ye\textsuperscript{1}, Bing Su\textsuperscript{2}, Ji-Rong Wen\textsuperscript{2}, Zheng Wang\textsuperscript{1}\\
\textsuperscript{1}Alibaba Group\,\raisebox{-3pt}{\includegraphics[height=1.2em]{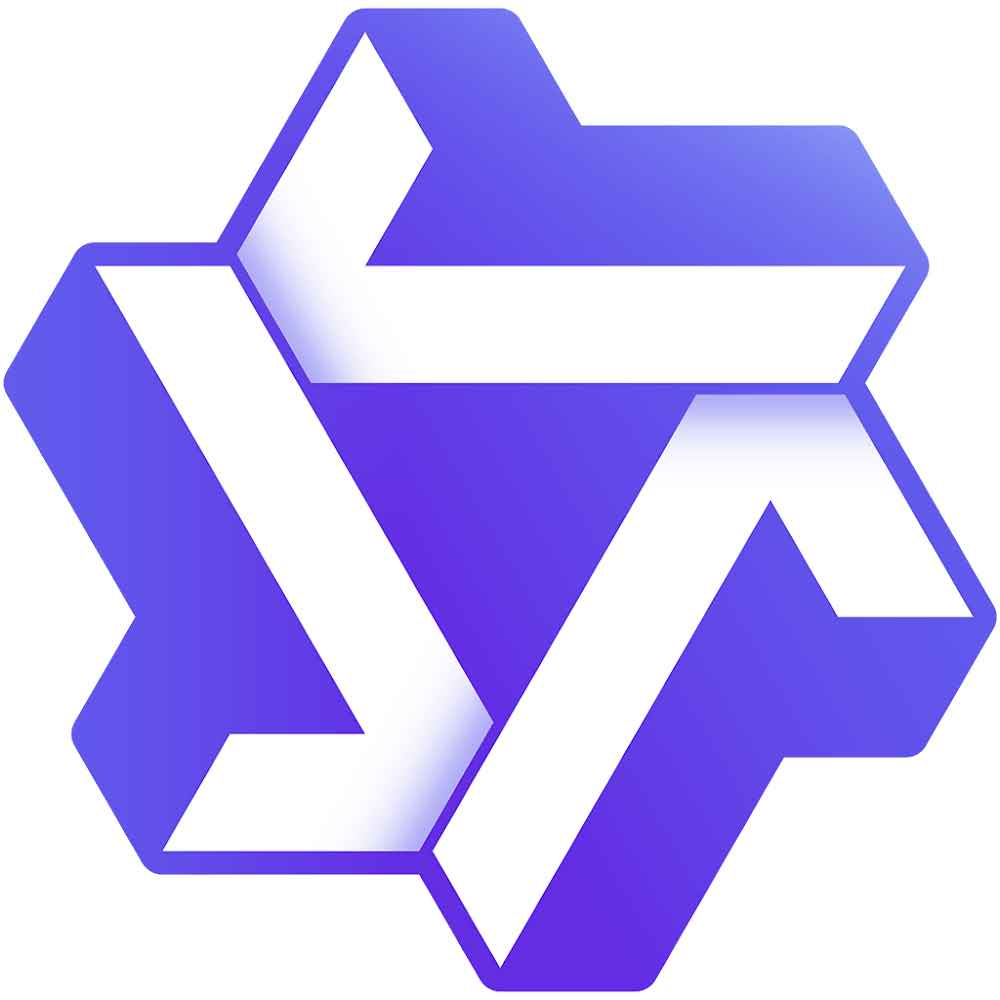}}\\
\textsuperscript{2}Gaoling School of Artificial Intelligence, Renmin University of China\,\raisebox{-3pt}{\includegraphics[height=1.2em]{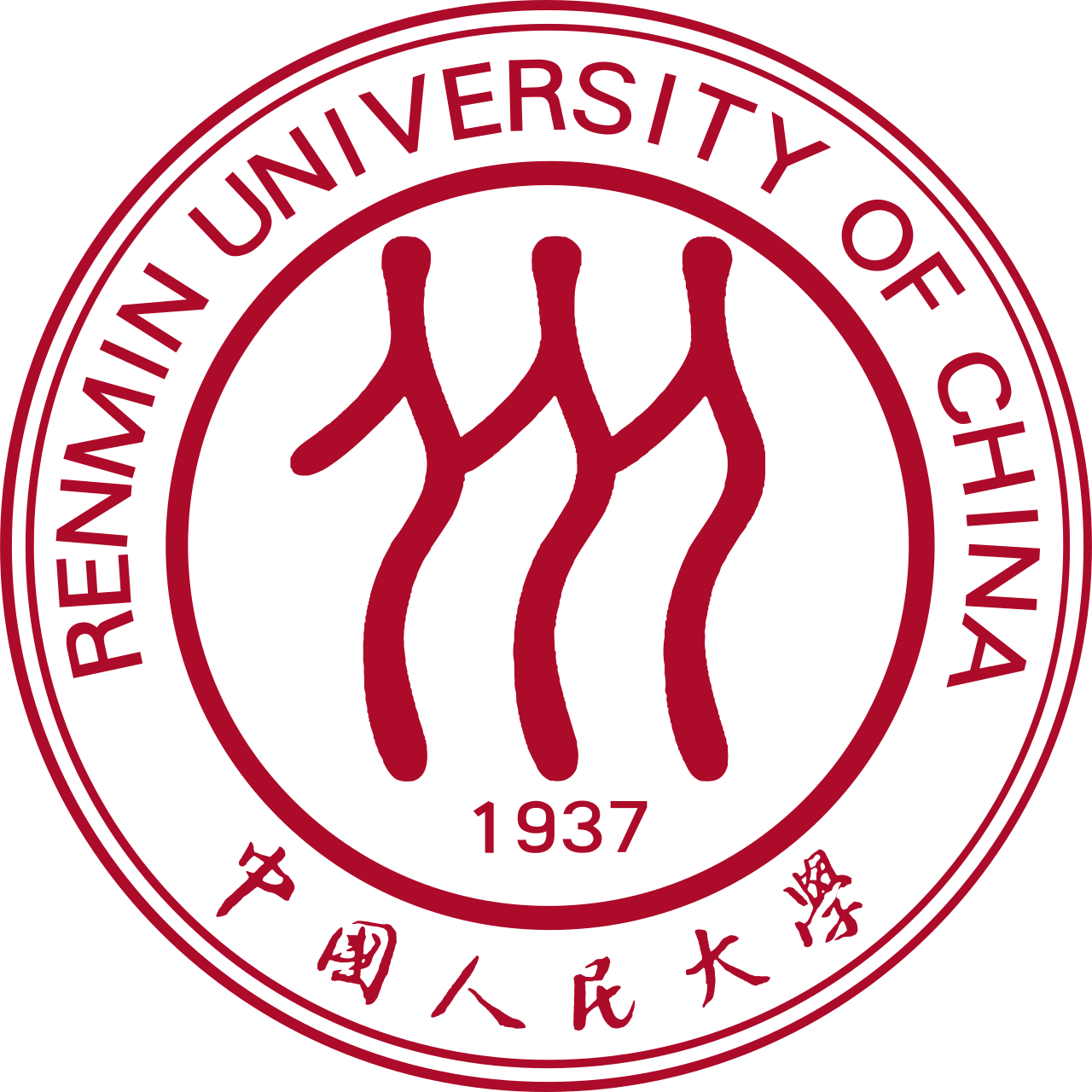}}
}

\newcommand{\OURS}{\textbf{LOGOS}\xspace}

\newcommand{\ignore}[1]{}

\begin{document}

\maketitle

\blfootnote{\noindent\footnotesize
\textsuperscript{*}Core contributors. \quad
}

\begin{abstract}
In this report, we present \textbf{LOGOS} (\textbf{L}anguage \textbf{O}f \textbf{G}enerative \textbf{O}bjects in \textbf{S}cience), a scientific generative language model that unifies heterogeneous tasks across the natural sciences within a single autoregressive
  framework based on a shared scientific grammar. It encodes diverse scientific objects and their three-dimensional interactions as token sequences over a common vocabulary. By representing spatial contact and
  constraint patterns as discrete tokens, the model captures complex structural interactions in a purely sequential manner, without relying on explicit coordinates or geometric neural networks. This unified
  representation enables a wide range of downstream tasks to be formulated consistently as next-token prediction in the same grammar space, creating strong alignment between continued multi-domain pre-training and
  downstream objectives. Across diverse tasks, \textbf{LOGOS} consistently matches or outperforms domain-specific baselines, providing preliminary evidence for the feasibility of "one model fits all" in the natural
  sciences. We train LOGOS models at different scales (1B, 3B, and 8B parameters) and find a consistent positive correlation between model size and performance. This suggests that the future of AI for Science (AI4S) may not lie in building an independent technical stack that is separated from large language models (LLMs). Instead, it may depend on deeply aligning scientific foundation models with LLMs through shared architectures, shared training paradigms, and shared inference infrastructure, so that LLMs can truly become a new entry point for AI4S. We release the model weights and associated resources to facilitate further research.

\end{abstract}

% \vspace{0.5cm}
\begin{figure}[h]
    \centering
    \includegraphics[width=1.0\linewidth]{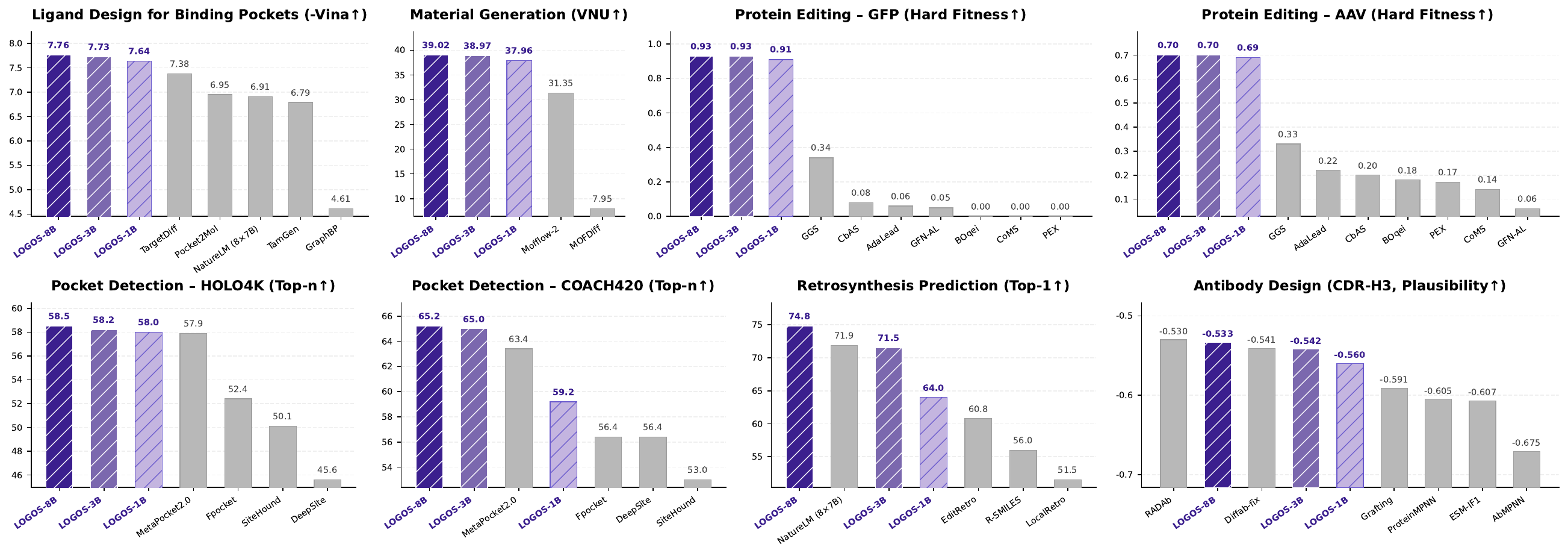}
    \caption{Benchmark performance of \OURS.}
    \label{fig:abs_fig}
\end{figure}
\vfill

\clearpage
\makeatletter
\def\addcontentsline#1#2#3{%
  \addtocontents{#1}{\protect\contentsline{#2}{#3}{\thepage}{\@currentHref}}%
}
\makeatother
\setcounter{tocdepth}{3}
\tableofcontents
\clearpage

\section{Introduction}

In recent years, the rapid advancement of artificial intelligence has fundamentally transformed the paradigm of scientific research, exerting an increasing influence across fields such as drug discovery, protein engineering, chemical synthesis, and materials design~\citep{abramson2024accurate,hayes2025simulating,m2024augmenting,zeni2025generative}. Within these domains, the goal extends beyond understanding and prediction to the generation of new entities subject to specific objectives and constraints. Whether discovering candidate ligands, designing functional materials, or planning reaction pathways, the core challenge is not merely selecting the best option from a finite set, but instead proposing novel candidates that satisfy multiple requirements within complex and heterogeneous design spaces~\citep{sanchez2018inverse, wang2023scientific}.
Consequently, the generative and creative capabilities of artificial intelligence provide a critical foundation for transforming scientific research into a design- and discovery-centered paradigm.

However, most previous efforts have primarily focused on representation learning for understanding and prediction. A representative example is the pre-training and fine-tuning paradigm based on BERT~\citep{DBLP:conf/naacl/DevlinCLT19}, in which models are pretrained on large-scale unlabeled data using self-supervised objectives, such as masked language modeling and contrastive learning, and then adapted to downstream tasks via fine-tuning. Under this paradigm, studies including ESM series~\citep{rives2021biological,lin2023evolutionary,hayes2025simulating}, BiT~\citep{DBLP:journals/corr/abs-2503-04362}, UniMol~\citep{DBLP:conf/iclr/ZhouGDZXWZK23}, DrugCLIP~\citep{jia2026deep}, and LucaOne~\citep{he2025generalized} have achieved substantial progress across diverse domains of the life sciences. Despite these advances, this paradigm entails inherent limitations. On the one hand, pretraining objectives such as masked reconstruction or contrastive alignment often exhibit substantial semantic misalignment with real downstream tasks. While these objectives focus on learning general recoverable or discriminative representations, downstream tasks require optimization toward more complex, task-specific goals such as binding affinity or synthetic accessibility. As a result, pretrained representations often lack direct in-context transferability and usually require targeted fine-tuning with downstream supervision to recalibrate the representation space. On the other hand, encoder-only architectures generally lack a unified native mechanism for conditional generation and thus typically depend on auxiliary generative modules or carefully designed sampling strategies to support generative tasks.

In light of the above limitations, generative modeling based on autoregressive language models offers a promising alternative. By explicitly modeling the joint probability distribution of sequences and learning conditional dependencies through next-token prediction, such approach naturally aligns the generation process with the training objective. In recent years, studies across diverse biological domains, including proteins, nucleic acids, and genomes (e.g., ProGen2~\citep{nijkamp2023progen2}, ProtGPT2~\citep{ferruz2022protgpt2}, Evo series~\citep{nguyen2024sequence,brixi2026genome}, GENERator~\citep{wu2025generator}), have demonstrated the potential of this paradigm. However, most existing generative models remain restricted to a single biological domain and lack a unified framework for cross-domain modeling. As a result, they struggle to capture shared principles and synergistic relationships across different biological modalities. Given that many biological processes involve complex interactions among proteins, small molecules, and other biomolecules, single-domain generative frameworks are often insufficient for modeling such complexity. Therefore, developing a unified multi-domain generative framework is of significant importance.

Toward this goal, drawing inspiration from the unified generative task modeling approach of general-purpose large language models, NatureLM~\citep{xia2025nature} attempts to use natural language as a universal interface for task description, recasting cross-domain heterogeneous biological tasks as conditional generation problems in an "instruction–response" format. Although this paradigm represents an important step toward building a general-purpose model for the natural sciences, its reliance on natural language as the core intermediary representation introduces several potential limitations. First, compared to the extremely abundant training corpora available for natural language, data in the biological and chemical domains remain considerably limited.
Under such imbalanced data regimes, adopting natural language as the cross-task interface may cause the model to disproportionately leverage generic linguistic patterns, at the expense of adequately capturing the structural regularities intrinsic to domain-native modalities and the associations that span across domains.
Second, natural scientific knowledge is typically not conveyed primarily through natural language, but is instead embodied in domain-specific representational formats, such as SMILES strings for small molecules~\citep{weininger1988smiles} and amino acid sequences for proteins~\citep{comm1968one}. These representations differ substantially from natural language in terms of compositional rules, structural constraints, and semantic mechanisms. Consequently, transferring general-purpose large language models to biological or chemical tasks inevitably confronts a non-trivial modality gap~\citep{hart2026protein,xian2025molrag}. Furthermore, under a fixed parameter budget, model capacity must be allocated and traded-off among competing capabilities~\citep{hernandez2021scaling}. Devoting a substantial share of this capacity to preserving broad natural language understanding and generation abilities may yield only limited marginal gains on domain-native tasks in the natural sciences. Prioritizing model capacity for specialized representations can be more parameter-efficient in tasks driven primarily by domain-native modalities.

We therefore approach the problem directly from the native structures of scientific objects and
their spatial interaction principles, recasting multi-domain scientific tasks as a unified generative token-modeling problem. The core insight underlying this perspective is that, although proteins, small molecules, materials, and reaction systems exhibit heterogeneity at the level of symbolic representation, they all fundamentally obey specific compositional rules, structural constraints, and interaction semantics, and can thus be viewed as instances of a shared "scientific language."
A unified formal grammar can map downstream problems onto sequence prediction under a common generative paradigm, which would enable cross-domain knowledge transfer, synergistic multi-task optimization, and close alignment between pre-training and downstream objectives within a single model.

\begin{figure}[th]
    \centering
    \includegraphics[width=\textwidth]{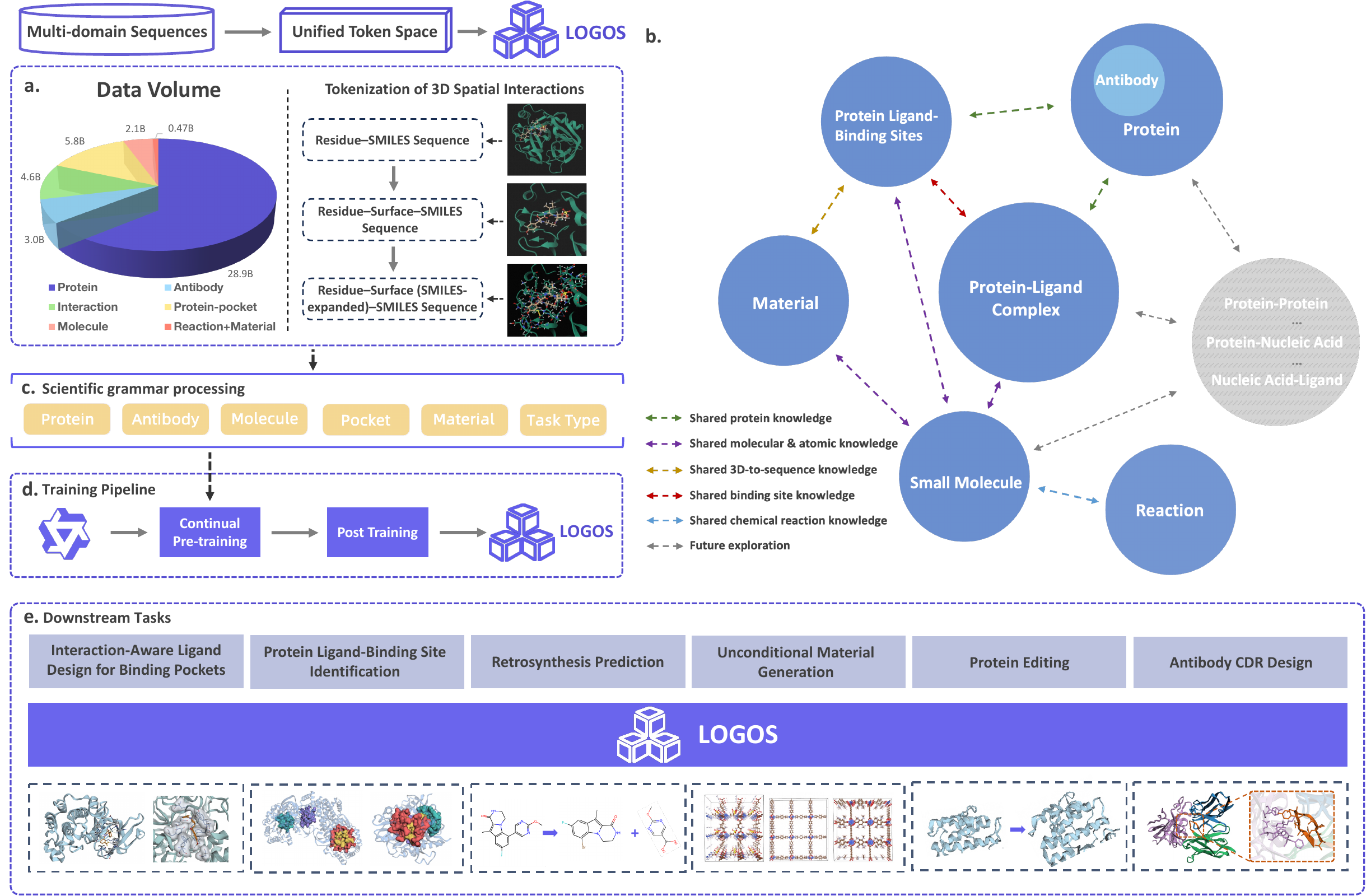}
    \caption{\textbf{Overview of LOGOS, a multi-domain generative framework grounded in a unified "scientific grammar".} LOGOS maps diverse scientific objects and their cross-object relationships into a shared discrete token space and unifies multi-domain tasks as autoregressive token modeling problems. Specifically, LOGOS is pretrained and post-trained with multi-domain scientific sequential data and interaction-aware relational sequences processed by the proposed scientific grammar, comprising a total of 44.87 billion tokens, based on a general large language model backbone. This training strategy unifies heterogeneous data across domains while bridging pretraining and downstream tasks, enabling joint learning of diverse scientific modalities within a single model. By leveraging the shared multidomain and multidimensional knowledge acquired during training, LOGOS achieves superior or competitive performance to specialized models across a broad range of downstream tasks, without explicit geometric dependencies. }
    \label{fig:main}
\end{figure}

To this end, as shown in Figure~\ref{fig:main}, we propose LOGOS, the first multi-domain generative framework based on a unified "scientific grammar." The framework uses this grammar as a shared representational interface across modalities, tasks, and hierarchical levels, encoding diverse scientific objects and their interactions as token sequences over a shared vocabulary. Through autoregressive continued pre-training, the model learns the joint distribution over these sequences. As pre-training operates within a grammar space closely aligned with downstream tasks, the model directly learns \textit{how to generate scientific objects}, \textit{how to construct inter-object relationships}, and \textit{how to perform scientific design under constraints}. This design ensures formal consistency and improved objective alignment between pre-training and downstream applications. Notably, in contrast to most approaches that rely on explicit three-dimensional coordinates, geometric neural networks, or dedicated structural modules~\citep{krishna2024generalized,schneuing2024structure,passaro2025boltz,hu2025target}, we explore an alternative pathway. Rather than directly treating such structures as continuous geometric inputs, we discretize, grammaticalize, and tokenize the key relationships and interaction patterns of scientific objects, incorporating them into a unified sequence generation framework.
For example, for the interaction between protein pockets and ligands, instead of explicitly modelling atomic-level coordinates, we design sequential representations that capture spatial contact and constraint information, enabling the model to learn pocket–ligand interaction patterns in a language modeling fashion.
This design not only circumvents the inconsistent availability of three-dimensional structural data across different domains, but also facilitates a deep-level unification of heterogeneous multi-domain scientific data within a common grammar system.

Specifically, we construct multi-domain training data spanning proteins, antibodies, small molecules, chemical reactions, materials, and their interactions, all unified under the proposed scientific grammar. Meanwhile, we reformulate a series of diverse downstream scientific tasks as next token prediction problems, including conditional ligand generation, materials generation, retrosynthesis prediction, pocket identification from protein sequences, protein optimization editing, and antibody CDR region prediction. Through this approach, we bring multi-domain scientific objects into a shared representation, bridge the gap between pre-training and downstream tasks, and ultimately subsume understanding, prediction, and design under a single modeling paradigm. Together, these advances constitute a concrete step toward a genuinely general-purpose "one model fits all" framework for the natural sciences.

In summary, the main contributions of this work are as follows:

1. \textbf{We propose a new generative modeling paradigm for AI-driven natural science.} We unify tasks involving proteins, antibodies, small molecules, materials, reactions, and pockets as generative token modeling problems, advancing natural science tasks from modality-specific specialized modeling toward general-purpose generative modeling within a unified grammar space. This perspective ensures consistency between the model's training objective and the ultimate application scenarios of scientific design, providing a new methodological foundation for generation-driven scientific discovery.

2. \textbf{We construct LOGOS, the first multi-domain generative framework based on a unified "scientific grammar."} We design a unified "scientific grammar" that incorporates different scientific objects and their cross-object relationships into a common discrete token space. In particular, by grammaticalizing spatial interactions into token representations, we achieve sequential modeling of complex structural information such as protein pockets, ligands, and their mutual interactions. This design simultaneously unifies multi-domain data and bridges pre-training with downstream tasks, allowing heterogeneous scientific data to be jointly learned within a single model. Our practice suggests that the future of AI4S may not lie in building an independent technical stack that is separated from LLMs. Instead, it may depend on deeply aligning scientific foundation models with LLMs through shared architectures, shared training paradigms, and shared inference infrastructure, so that LLMs can truly become a new entry point for AI4S.

3. \textbf{We achieve unified high-performance multi-domain, multi-task modeling without explicit geometric dependencies.} Working within a purely sequential paradigm, LOGOS achieves strong performance on a range of representative tasks, including conditional ligand generation given a target protein, materials generation and property-related tasks, retrosynthesis prediction, pocket mining, protein editing, and antibody CDR region prediction. Experimental results demonstrate that, even without relying on explicit 3D geometric networks, the framework can learn effective cross-modal structural regularities through the unified scientific grammar, and surpass domain-specific methods with a relatively lightweight model scale, validating the feasibility of "one model fits all" in the natural sciences.

4. \textbf{We open-source the model and associated resources to facilitate community research and development.} To support follow-up research, we release the model weights, training data processing pipelines, and related resources, aiming to provide a reusable technical pathway for building general-purpose scientific foundation models and to promote continued community progress in unified representation, unified generation, and cross-domain scientific intelligence.

\section{Method}
\subsection{Data Construction}

In the natural sciences, elucidating the molecular basis and emergent behaviors of living systems, as well as constructing and regulating complex functional chemical systems from molecular components, have become central research directions spanning the life sciences, chemistry, and their interdisciplinary intersections~\citep{kitano2002systems,lehn2009towards}. Within this context, proteins and small molecules represent two fundamental classes of molecular entities for characterizing biological activity and enabling chemical intervention, respectively. Proteins, as the primary executors of biological processes, participate in virtually all cellular biochemical events and constitute one of the most extensively studied subjects in modern biology and biomedical research~\citep{american2004molecular}. Small molecule compounds, in turn, serve as the most widely employed class of chemical agents for modulating protein function and intervening in biological processes, forming the material foundation of drug discovery and chemical biology~\citep{schreiber2015advancing}.

The knowledge system surrounding these two core molecular entities can be further expanded along two dimensions: molecular family expansion and inter-molecular interface interactions. Along the first dimension, antibodies—a functionally specialized subclass of proteins possessing unique combinatorial sequence diversity and programmable specificity—warrant dedicated treatment beyond generic protein representations~\citep{tonegawa1983somatic}; chemical reactions and functional materials further extend the representational scope of molecular chemistry beyond static small-molecule structures to encompass transformation processes and hierarchical compositions. Along the second dimension, the interface interactions between proteins and small molecules, particularly the selective recognition of small-molecule ligands by protein binding pockets, represent a critical link bridging biomacromolecules and chemical entities, and underpin our understanding of molecular recognition mechanisms and rational drug design~\citep{kitchen2004docking}.

Based on the above knowledge landscape centered on proteins and small molecules while encompassing their family expansions and interface interactions, as shown in Figure~\ref{fig:main}a and Figure~\ref{fig:main}b, we construct a large-scale pre-training corpus covering seven modalities: protein, antibody, small molecule, chemical reaction, materials, protein ligand-binding sites, and protein–ligand complex data. Among these, protein and antibody data form the biomacromolecular layer; small molecule, chemical reaction, and materials data form the chemical entity and transformation process layer; and protein ligand-binding sites and protein–ligand complex data explicitly capture the interaction interfaces bridging the two. The summary of the data construction can be found in Appendix~\ref{Overview of data construction}, and all seven modalities are encoded as discrete token sequences under a shared vocabulary within a unified scientific grammar framework, which is shown in Figure~\ref{fig:data construction}. The following subsections introduce the specific construction and grammar design for each modality of sequence data in turn.

\subsubsection{Protein}

Protein sequence data constitutes the foundational modality of the biomacromolecular knowledge layer. We adopt the UniRef90 dataset~\citep{suzek2015uniref} as the source of protein sequence data. UniRef90 is a non-redundant protein sequence clustering dataset provided by UniProt~\citep{ahmad2025uniprot}, which clusters UniProtKB entries at a 90\% sequence identity threshold, effectively reducing sequence redundancy in large-scale protein data while preserving good sequence coverage and biological diversity. As such, UniRef90 has become one of the standard data sources commonly used in protein representation learning and large-scale pre-training.

During data preprocessing, we extract protein sequences in entry order from the UniRef90 FASTA files and convert them into linear sequences composed of standard amino acid letters. To explicitly distinguish different scientific entity types within the shared vocabulary framework and enhance the model's ability to recognize modality boundaries, we prepend and append boundary tokens \texttt{<ProteinS>} and \texttt{<ProteinE>} to each protein sequence, thereby encoding the raw sequence into a discrete token sequence of the form \texttt{<ProteinS>...<ProteinE>}. Below is an example.

\begin{promptbox}
\texttt{<ProteinS>GSRSIMEYLCG<ProteinE>}
\end{promptbox}

\subsubsection{Antibody}

Antibodies are the principal effector molecules of the humoral immune system, mediating specific antigen recognition and downstream immune functions. They exhibit high sequence diversity, structural plasticity, and antigen-binding specificity, and play critical roles in immune defense against infection, disease diagnosis, therapeutic antibody development, and protein engineering~\citep{tonegawa1983somatic}. As a functionally specialized subclass of proteins, antibodies not only occupy a pivotal position in biomedical research, but also offer unique data resources for studying hypervariable sequence spaces, molecular recognition patterns, and principles of biomacromolecular design. Therefore, incorporating the antibody modality beyond general protein sequence data enriches the biomacromolecular knowledge layer with representations of immune recognition and specific binding.

The antibody data used in this study are sourced from the Observed Antibody Space (OAS) database~\citep{olsen2022observed}. OAS is a widely used large-scale repository that systematically aggregates sequences from publicly available immune repertoire sequencing studies. We extract unpaired data from OAS as the raw corpus. Subsequently, we remove sequences with unclear provenance or missing donor metadata, missing isotype information, or containing non-standard or ambiguous residues (e.g., X, B, Z), and further exclude entries lacking IMGT-defined framework region annotations, in order to ensure data consistency, interpretability, and accuracy. On this basis, to further reduce redundancy introduced by high-frequency clonally related sequences, we cluster the filtered antibody sequences at 70\% sequence identity and retain one representative sequence per cluster. For grammar representation, we encode each antibody sequence as an amino acid token sequence, with boundary tokens \texttt{<AntibodyS>} and \texttt{<AntibodyE>} prepended and appended, respectively, forming the representation \texttt{<AntibodyS>...<AntibodyE>}. Below is an example.

\begin{promptbox}
\texttt{<AntibodyS>QSALTQPASVSGSPGQSITISCTGSSSDVGGYNYVSWYQQHPDKAPKLMIYEVSNRPSGVSTRFSGSKSGN\\TASLTISGLQPEDEAAYYCSSYTTTKTLVFGGGTKLTVL<AntibodyE>}
\end{promptbox}

\subsubsection{Small Molecule}

Small molecule data constitute the foundational modality of the chemical entity knowledge layer, encoding information such as atomic composition, bond connectivity, stereochemistry, and molecular topology. Small molecules represent fundamental units for characterizing chemical space and linking molecular design to biological modulation~\citep{schreiber2015advancing}. Therefore, incorporating the small molecule modality into the multimodal scientific corpus is essential for enriching the chemical knowledge layer and establishing cross-modal associations with biomacromolecules.

We source our small molecule data from the large-scale quantum chemistry database PubChemQC B3LYP/6-31G//PM6~\citep{nakata2023pubchemqc}, which is derived from the PubChem molecular library. We extract molecular representations in SMILES format—a widely used linear notation encoding atom types, bond connectivity, aromaticity, and chirality~\citep{weininger1988smiles}—and perform structural consistency filtering based on the corresponding three-dimensional conformations, ensuring high structural fidelity of the retained molecules at both the topological and stereochemical levels. After filtering, all retained molecules are represented by their corresponding canonical SMILES sequences. To explicitly distinguish scientific entity types, we prepend and append boundary tokens \texttt{<MoleculeS>} and \texttt{<MoleculeE>} to each SMILES string, encoding it as a discrete token sequence of the form \texttt{<MoleculeS>...<MoleculeE>}. Below is an example.

\begin{promptbox}
\texttt{<MoleculeS>O=C(Nc1cccc2cn[nH]c12)c1ccc(Br)s1<MoleculeE>}
\end{promptbox}

\subsubsection{Reaction}

Chemical reactions are the core processes underlying material transformation and synthetic pathway construction, and serve as a critical bridge connecting the known compound space with the design of novel functional molecules. Modeling reaction processes not only facilitates understanding of the intrinsic rules governing molecular transformations, but also directly supports applications such as retrosynthetic planning, reaction condition optimization, and yield prediction~\citep{deng2025rsgpt}. Within the knowledge landscape described above, the chemical reaction modality extends the small-molecule chemical space by supplementing information on intermolecular relationships and transformation pathways from the perspective of dynamic processes, complementing static small-molecule structural data to jointly form the chemical entity and transformation process knowledge layer.

The chemical reaction data used in this study are sourced from two large-scale open reaction databases: the Open Reaction Database (ORD)~\citep{kearnes2021open} and ECReact~\citep{probst2022biocatalysed}. ORD systematically collects experimentally verified reaction records in organic chemistry; Ecreact focuses on enzyme-catalyzed reactions, providing diverse biochemical transformation data. During data preprocessing, we select reaction entries from these two datasets for which canonical SMILES representations of both reactants and products can be successfully extracted, and discard entries with incomplete structural information or unparseable SMILES, to ensure the accuracy and usability of each reaction record at the chemical topology level. In terms of grammar design, the representation of chemical reactions builds upon the small molecule grammar. Specifically, each participating molecule in a reaction, whether a reactant or a product, reuses the \texttt{<MoleculeS>...<MoleculeE>} boundary tokens defined in the small molecule modality, thereby maintaining cross-modal consistency of chemical entity representations at the token level. On this basis, we introduce directional tokens \texttt{<React>} and \texttt{<ReverseReact>} to explicitly encode the direction of transformation: \texttt{<React>} denotes the forward reaction from reactants to products, while \texttt{<ReverseReact>} denotes the reverse derivation from products to reactants. Accordingly, each chemical reaction is bidirectionally sampled into forward and reverse training sequences, with the general forms \texttt{<MoleculeS>...<MoleculeE><React><MoleculeS>...<MoleculeE>} and \texttt{<MoleculeS>...<MoleculeE><ReverseReact><MoleculeS>...<MoleculeE>}, respectively. Below is an example.

\begin{promptbox}
\texttt{<MoleculeS>CC(C)(C)OC(=O)NNC(=O)c1ccc(Br)cc1Cl<MoleculeE><ReverseReact>\\<MoleculeS>O=C(O)c1ccc(Br)cc1Cl<MoleculeE><MoleculeS>CC(C)(C)OC(=O)NN<MoleculeE>}
\end{promptbox}

\subsubsection{Material}

Functional materials extend the representational scope of molecular chemistry along the dimension of hierarchical composition, bridging molecular-level chemistry and the construction of macroscopic functional systems. In this work, we focus on Metal-Organic Frameworks (MOFs) as a representative class of functional materials. MOFs are crystalline porous materials formed through the self-assembly of inorganic metal clusters (nodes) and organic linkers via coordination bonds. Owing to their structural tunability, high porosity, and broad application prospects, MOFs have emerged as a prominent class of designable porous materials in materials science~\citep{kim2026flexible}.

The materials data used in this study are sourced from the large-scale hypothetical MOF structure database constructed by~\citet{boyd2019data}. From a compositional perspective, MOFs possess a characteristic dual-component architecture—where the organic linkers themselves are small-molecule compounds with well-defined topologies and bond connectivity, while the metal clusters define the coordination environment and connection topology of the framework. During data preprocessing, we employ a structural decomposition strategy based on metal-oxo decomposition~\citep{kim2026flexible} to systematically decompose each MOF structure into its metal cluster and organic linker components. On this basis, we further validate the chemical integrity of the organic linkers: only those parseable into valid molecular graphs using RDKit~\citep{landrum2025rdkit} are retained, while entries containing unparseable fragments, broken bonds, or non-standard chemical structures are discarded, ensuring that every retained organic linker is a chemically valid molecular entity. In terms of grammar design, the representation of the materials modality maintains consistency with the unified grammar framework. Specifically, we use \texttt{<MaterialS>} and \texttt{<MaterialE>} as the outer boundary tokens of the entire material sequence to delimit the start and end of a material entity. Within the material sequence, metal cluster components are wrapped with dedicated boundary tokens \texttt{<MetalS>} and \texttt{<MetalE>}, while organic linker components directly reuse the \texttt{<MoleculeS>} and \texttt{<MoleculeE>} boundary tokens defined in the small molecule modality. In this way, a complete material data entry is represented as the following nested structure: \texttt{<MaterialS><MetalS>...<MetalE>...<MoleculeS>...<MoleculeE>...<MaterialE>}. Below is an example.

\begin{promptbox}
\texttt{<MaterialS><MetalS>[Cu+][Cu+]<MetalE><MoleculeS>CO[N-][N-]OC<MoleculeE>\\<MoleculeS>O=C([O-])/C(F)=C(\textbackslash \textbackslash F)C(=O)[O-]<MoleculeE><MoleculeS>CO[C]([C](OC)[C](F)C(=O)[O-])[C](F)C(=O)[O-]<MoleculeE><MaterialE>}
\end{promptbox}

\subsubsection{Protein Ligand-Binding Sites}

Protein ligand-binding sites, commonly referred to as binding pockets, are localized spatial regions within the three-dimensional structure of a protein responsible for specific interactions with small-molecule ligands~\citep{gao2013comprehensive}. Accurate identification and characterization of binding pockets is not only a key prerequisite for understanding protein–ligand molecular recognition mechanisms, but also forms the foundation for structure-based drug design tasks such as virtual screening and lead compound optimization~\citep{isert2023structure}. Within the knowledge landscape described above, the protein pocket modality serves as a bridge connecting the biomacromolecular knowledge layer and the chemical entity knowledge layer—on one hand, a pocket is a local substructure of a protein sequence; on the other hand, its functional role lies in binding with small-molecule ligands. It therefore naturally resides at the interface between proteins and small molecules, the two core molecular entities. Incorporating pocket information into the pre-training corpus enables the model to explicitly learn associations between protein sequence context and ligand-binding function within a unified grammar space, directly benefiting downstream tasks such as conditional ligand generation and binding site prediction.

The protein pocket data used in this study are derived from protein structures in the Protein Data Bank (PDB)~\citep{berman2000protein}. We employ P2Rank~\citep{krivak2018p2rank} to predict binding sites on each protein structure. P2Rank is a machine-learning-based method capable of scoring and ranking potential binding regions on the protein surface with high accuracy. For each protein sample, we select the top-10 ranked pockets based on P2Rank confidence scores and map the residue positions of each pocket back to the protein's primary sequence, thereby obtaining the amino acid composition and positional information of each pocket at the sequence level.

\textbf{Tokenization of Spatial Interaction.} To establish connections among scientific modalities across multiple domains, we aim to represent interactions among different scientific entities in order to facilitate knowledge integration across heterogeneous scientific data. To this end, we design a tokenization method for spatial interactions, which represents the relationships among different scientific entities within a purely sequential framework. A protein ligand-binding pocket is fundamentally a three-dimensional structural entity: it is determined primarily by the spatial proximity of residues that collectively shape the ligand-recognition interface, rather than by their adjacency in the primary sequence. Tokenizing such pockets therefore provides a natural instantiation of our central principle—encoding spatial interactions between biological entities into a unified discrete grammar. The four representations below follow a deliberately progressive design, each exposing the protein–ligand interface at a finer granularity. The first representation localizes the interaction by marking, within the linear sequence, those residues defined as belonging to the pocket under a given structural criterion, thereby projecting the 3D interface onto a one-dimensional token stream. Building on this, because pocket–ligand interactions are largely mediated by side-chain chemistry, the second representation expands each pocket residue into a predefined SMILES-based proxy, placing the interface description in the same chemical representation space as the ligand and introducing an explicit, representation-level correspondence between amino acids and small-molecule fragments. The third representation makes this cross-modal mapping itself a learning target, directly modeling the transformation between the amino-acid sequence form and the SMILES-expanded pocket form. Finally, the fourth representation aligns the formulation with the downstream task of identifying potential binding sites from protein sequence alone, casting interface discovery as a generative sequence-prediction problem. Together, these four representations progress from \emph{locating} the spatial interaction, to \emph{chemically characterizing} it, to \emph{learning its cross-representation mapping}, and finally to \emph{predicting} it—collectively illustrating how key aspects of a spatial interaction can be systematically encoded within a unified grammar.

In terms of grammar design, to characterize protein pocket information from multiple complementary perspectives, we design four sequence representations for each pocket of every protein sample:

\textbf{(1) Amino acid-level pocket annotation sequence.} This representation explicitly annotates the positional information of pocket residues within the protein sequence. Specifically, within the linear amino acid sequence of a protein, the segments corresponding to a pocket are annotated in place using \texttt{<PocketS>} and \texttt{<PocketE>}, while the entire sequence retains \texttt{<ProteinS>} and \texttt{<ProteinE>} as outer boundary tokens. This allows the model to localize the pocket within the global sequence context. Below is an example.

\begin{promptbox}
\texttt{<ProteinS>HMIS...<PocketS>A<PocketE>DLA...<PocketS>N<PocketE>SHS...<ProteinE>}
\end{promptbox}

\textbf{(2) Small-molecule-expanded pocket sequence.} The side-chain groups of pocket residues typically serve as the primary structural basis for physicochemical interactions with ligands. To establish an explicit link between pocket residues and the small-molecule chemical space, this representation expands each pocket residue into its corresponding side-chain SMILES. The content enclosed by \texttt{<PocketS>...<PocketE>} thus becomes a SMILES string rather than a single amino acid letter, bridging the protein and small-molecule representation spaces at the token level. Below is an example.

\begin{promptbox}
\texttt{<ProteinS>HMIS...<PocketS>C[C@@H](C(=O)O)N<PocketE>DLA...<PocketS>C([C@@H](C(=O)O)N)C(=O)N<PocketE>SHS...<ProteinE>}
\end{promptbox}

\textbf{(3) Amino acid–small molecule transformation sequence.} This representation explicitly aligns the above two forms by introducing a directional token \texttt{<Trans>} to concatenate the amino acid-level sequence and the SMILES-expanded sequence into a single transformation sequence. This enables the model to directly learn the mapping between amino acid identifiers and their corresponding molecular structures during autoregressive generation. Its general form is: \texttt{<ProteinS>...<PocketS>amino acid<PocketE>...<ProteinE><Trans><ProteinS>...<PocketS>SMILES<PocketE>...<ProteinE>}. Below is an example.

\begin{promptbox}
\texttt{<ProteinS>HMIS...<PocketS>A<PocketE>DLA...<PocketS>N<PocketE>SHS...<ProteinE><Trans><ProteinS>HMIS...<PocketS>C[C@@H](C(=O)O)N<PocketE>DLA...<PocketS>C([C@@H](C(=O)O)N)C(=O)N<PocketE>SHS...<ProteinE>}
\end{promptbox}

\textbf{(4) Binding site identification sequence.} Predicting potential binding sites from a protein sequence is a key downstream task in drug design~\citep{gao2013comprehensive}. To incorporate this task form during pre-training, we introduce a task semantic token \texttt{<Search>}, encoding the pocket identification process as a generative sequence prediction problem. A complete protein sequence without pocket annotations serves as the input condition, followed by \texttt{<Search>}, and then a pocket-annotated sequence is generated as the output target. This ensures that the pocket identification objective is inherently consistent with the autoregressive generation paradigm. Below is an example.

\begin{promptbox}
\texttt{<ProteinS>HMIS...<ProteinE><Search><ProteinS>HMIS...<PocketS>A<PocketE>DLA...<PocketS>N<PocketE>SHS...<ProteinE>}
\end{promptbox}

Together, these four representations provide complementary views of protein pockets—from sequence-level positional annotation to cross-modal structural alignment—within a single unified grammar. Notably, pocket tokens \texttt{<PocketS>...<PocketE>} are nested within the protein grammar \texttt{<ProteinS>...<ProteinE>}, and the pocket content can in turn be expanded into small-molecule SMILES, achieving cross-level grammar nesting that positions protein pockets as a grammatical hub connecting the protein sequence space and the small-molecule chemical space.

\subsubsection{Protein-Ligand Complex}

As described in the previous subsection, the protein pocket modality establishes a grammatical bridge between protein sequences and the small-molecule chemical space from the perspectives of binding site identification and multi-level pocket characterization. However, the characterization of the pocket itself does not yet address its true functional context—namely, the binding relationship between a pocket and a specific small-molecule ligand. The protein–ligand complex, as the structural embodiment of protein–small molecule interface interactions, explicitly encodes the core information of "which pocket residues bind to which ligand," and constitutes a key source of structural information for understanding molecular recognition specificity, binding affinity, and drug action mechanisms~\citep{du2016insights, sedov2025protein}. From the perspective of downstream applications, generating small-molecule ligands that can specifically bind to a given protein binding pocket—i.e., structure-based de novo drug design—is among the most practically relevant tasks in drug discovery~\citep{peng2022pocket2mol, schneuing2024structure}. Therefore, incorporating protein–ligand complex data on top of the pocket modality completes the protein–small molecule interface interaction knowledge layer. It also ensures that the sequence generation objective during pre-training aligns directly with conditional ligand generation, a critical downstream task.

The protein–ligand complex data used in this study are sourced from the Q-BioLiP dataset~\citep{wei2024q}. During data preprocessing, for each protein–ligand complex record, we perform distance-based selection of amino acid residues in the protein structure using radii ranging from 5 to 10 \AA{} centered on each atom of the ligand molecule. Residues falling within the distance threshold are identified as pocket residues and mapped back to the protein primary sequence. The use of multiple distance thresholds rather than a single fixed radius captures pocket–ligand interactions at multiple spatial scales: smaller radii focus on core residues in direct contact with the ligand and short-range non-covalent interactions, while larger radii further incorporate peripheral residues that influence pocket shape, electrostatic microenvironment, and solvent accessibility. Additionally, this multi-granularity sampling strategy generates multiple training sequences with varying spatial extents for each complex, enhancing the model's robustness to pocket boundary definitions.

In terms of grammar design, the representation of protein–ligand complexes integrates the grammars of the protein pocket modality and the small molecule modality. We adopt a molecular structure-level joint pocket–ligand representation: first, pocket residue positions are marked within the protein sequence using \texttt{<PocketS>} and \texttt{<PocketE>}, and the amino acid residues within the pocket are expanded into their corresponding small-molecule SMILES representations, thereby characterizing the chemical features of pocket residues at the molecular structure level; subsequently, the SMILES representation of the ligand is appended at the end of the protein sequence using the small molecule grammar \texttt{<MoleculeS>...<MoleculeE>}. In this way, a complete protein–ligand complex data entry is represented in the following form: \texttt{<ProteinS>...<PocketS>SMILES<PocketE>...<ProteinE><MoleculeS>SMILES<MoleculeE>}. 

Echoing the pocket representations introduced above, this joint protein–ligand sequence constitutes the complete tokenization of spatial interactions within our grammar. Specifically, distance-based, multi-radius residue selection translates continuous 3D proximity into a discrete, token-level record of the interface; concurrently, appending the ligand SMILES explicitly couples this interface to its cognate molecule. Together, they encode the complete “which residues bind which ligand” relationship in a single sequence. This joint representation directly extends the preceding pocket tokenization: it inherits both residue-level localization and SMILES-based chemical alignment, but crucially grounds them in an explicit protein–ligand pair. This completes the progression from characterizing an interaction interface in isolation to encoding a concrete spatial interaction in full. Below is an example.

\begin{promptbox}
\texttt{<ProteinS>NED...<PocketS>CC(C)C[C@@H](C(=O)O)N<PocketE>ALE...<PocketS>C[C@@H](C(=O)O)N<PocketE>VGF...<ProteinE><MoleculeS>OCCO<MoleculeE>}
\end{promptbox}

\begin{figure}[t]
    \centering
    \begin{subfigure}[b]{0.45\linewidth}
        \centering
        \vspace{-0.1cm}
        \includegraphics[width=\linewidth]{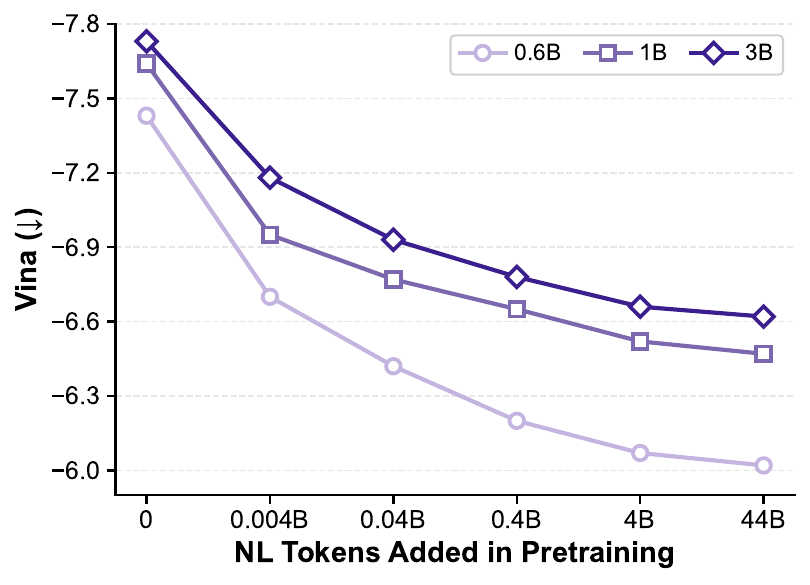}
        \caption{}
        \label{fig:nl_tokens}
    \end{subfigure}
    \begin{subfigure}[b]{0.48\linewidth}
        \centering
        \vspace{-0.1cm}
        \includegraphics[width=\linewidth]{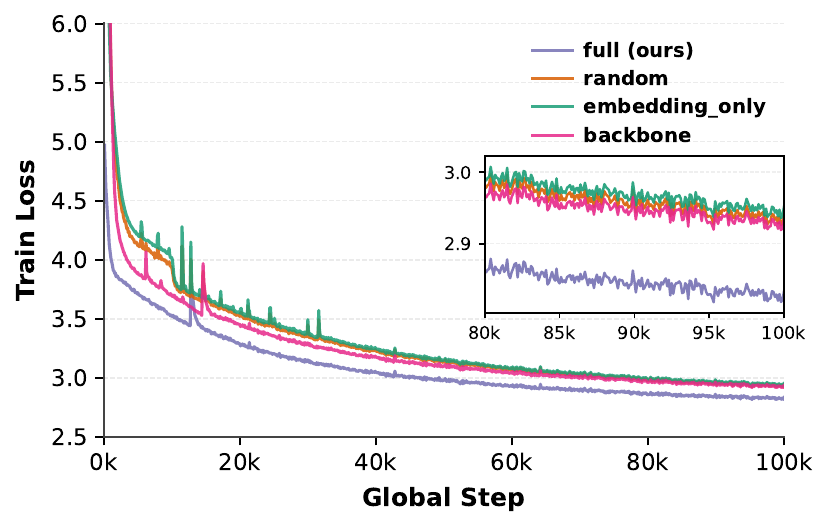}
        \caption{}
        \label{fig:loss_curve}
    \end{subfigure}
    \caption{\textbf{a.} Effect of natural language (NL) token volume during pretraining on ligand design performance across different model scales. Increasing NL tokens in pretraining consistently degrades Vina docking performance across model scales, indicating a capacity trade-off between general linguistic and domain-native capabilities under fixed parameter budgets. \textbf{b.} Pre-training loss curves for four weight initialization configurations defined in Table~\ref{tab:model components}. Configurations inheriting the Backbone (settings iii and iv) converge significantly faster and reach lower final loss, while inheriting only the Embedding (setting ii) yields even higher loss than full random initialization (setting i), consistent with the downstream performance trends in Table~\ref{tab:model components}.}
\vspace{-0.3cm}
\label{fig:ablation_nl}
\end{figure}

\subsection{Exploratory Studies}

Having completed the construction of the multimodal scientific corpus and the unified grammar design described above, this section reports a set of exploratory experiments aimed at providing empirical evidence and intuitive understanding to inform the design decisions of the framework and training pipeline, as well as offering valuable reference for future work in the domain community. These experiments are organized around three core questions: (1) the effect of pre-trained weights and natural language priors from large language models on learning scientific modalities; (2) the role of the unified scientific grammar in cross-modal knowledge alignment; and (3) whether cross-domain synergistic gains exist among multi-domain scientific data. The following subsections discuss each of these questions in turn.

\subsubsection{On the Role of Initialization Strategy and Linguistic Priors}

Generative scientific modeling based on autoregressive language models faces a fundamental architectural choice: should one inherit pre-trained weights from general-purpose large language models (LLMs)? What role do LLM pre-trained weights play in learning scientific modalities? To answer these questions, we use Qwen3-0.6B~\citep{yang2025qwen3} as the base model and conduct explorations from two perspectives: weight initialization strategy and natural language corpus ratio.

\begin{wraptable}[11]{r}{0.6\textwidth}
    \centering
    \caption{Ablation study on weight initialization strategies for ligand design. \ding{51} and \ding{55} indicate whether each component (Embedding, Backbone, LM\_head) is initialized from pre-trained LLM weights or randomly initialized, respectively.}
    \label{tab:model components}
    \small
    \begin{tabular}{@{}l|cccc@{}}
        \toprule
        & \textbf{Embedding} & \textbf{Backbone} & \textbf{LM\_head} & \textbf{Vina}$\downarrow$ \\
        \midrule
        random
        & \textcolor{red!70!black}{\ding{55}}
        & \textcolor{red!70!black}{\ding{55}}
        & \textcolor{red!70!black}{\ding{55}}
        & $-6.91$ \\

        embedding\_only
        & \textcolor{green!70!black}{\ding{51}}
        & \textcolor{red!70!black}{\ding{55}}
        & \textcolor{red!70!black}{\ding{55}}
        & $-6.78$ \\

        backbone
        & \textcolor{green!70!black}{\ding{51}}
        & \textcolor{green!70!black}{\ding{51}}
        & \textcolor{red!70!black}{\ding{55}}
        & $-7.21$ \\

        \midrule

        \textbf{full (ours)}
        & \textcolor{green!70!black}{\ding{51}}
        & \textcolor{green!70!black}{\ding{51}}
        & \textcolor{green!70!black}{\ding{51}}
        & $\mathbf{-7.43}$ \\
        \bottomrule
    \end{tabular}
\end{wraptable}

\textbf{Weight initialization.} The trainable parameters of the model consist of three main components: the embedding layer, the Transformer backbone, and the language modeling head (LM\_head). We design four initialization configurations by progressively inheriting LLM pre-trained weights for each component, as shown in Table~\ref{tab:model components}: (i) all three components randomly initialized; (ii) only the embedding initialized from LLM weights; (iii) both the embedding and backbone initialized from LLM weights; and (iv) all three components initialized from LLM weights. We evaluate performance on the key task of pocket-conditioned ligand generation. As shown in Table~\ref{tab:model components}, the configuration in which all components inherit LLM pre-trained weights (setting iv) achieves the best performance. Notably, inheriting only the embedding layer (setting ii) slightly degrades performance compared to full random initialization (setting i), whereas introducing the Backbone (setting iii) yields a substantial improvement. This suggests that the transferable benefits primarily reside in the Transformer's deep sequence modeling layers—including long-range dependency capture, contextual conditional inference, and hierarchical representation construction—rather than in the input token representations alone. The pre-training loss curves in Figure~\ref{fig:loss_curve} further corroborate this finding: configurations inheriting the Backbone converge significantly faster and reach lower final loss values, whereas inheriting only the embedding provides negligible acceleration. In other words, the compositional logic and sequential reasoning patterns embedded in natural language share, to some extent, abstract structural commonalities with the structural dependency patterns in scientific modalities, thereby providing beneficial inductive biases for scientific sequence modeling.

\textbf{Natural language corpus ratio.} Having confirmed the benefit of LLM weight initialization, we further explore the effect of mixing different proportions of general natural language corpus during the continued pre-training stage. As shown in Figure~\ref{fig:nl_tokens}, when we progressively increase the proportion of general natural language training data in the scientific corpus, model performance on the ligand generation task exhibits a consistent decline across all three model scales (0.6B, 1B, and 3B). This result points to an unavoidable capacity allocation trade-off under a fixed parameter budget: although LLM pre-trained weights serve as a beneficial initialization starting point for scientific modality learning, if the model is simultaneously required to maintain broad natural language understanding and generation capabilities during subsequent continued pre-training, this inevitably occupies model capacity that would otherwise be used for encoding domain-native structural regularities, thereby weakening its ability to finely model scientific modality entities.

The results of the above two sets of experiments jointly reveal a finding with important implications: the sequence modeling priors embedded in natural language pre-training can provide potential benefits for scientific modality learning at the weight level,
but directly introducing large amounts of natural language corpus during continued training instead has a negative impact on scientific modality learning at a comparable parameter scale. Therefore, rather than adopting natural language as a unified cross-modal interface, we choose to start directly from the native representations of scientific objects, using a unified scientific grammar rather than natural language as the cross-modal interface, and concentrating model capacity on native representation learning of scientific modalities, which achieves higher parameter efficiency in tasks driven by domain-native modalities.

\subsubsection{Validation of Unified Scientific Grammar}

\begin{table*}[tb]
\centering
\caption{\textbf{Ablation study on pretraining data composition.}}
\label{tab:ablation_pretrain}
\small
\begin{tabular}{@{}lccc@{}}
\toprule
\textbf{Method} & \textbf{Site Acc.}$\uparrow$ & \textbf{Sample Acc.}$\uparrow$ & \textbf{Vina}$\downarrow$ \\
\midrule
\OURS\mbox{-}1B w/o protein ligand\mbox{-}binding sites \& protein\mbox{-}ligand complex & 3.76  & 5.33  & $-$3.57 \\
\OURS\mbox{-}1B w/o protein ligand\mbox{-}binding sites\mbox{-}\textlangle Trans\textrangle                   & 17.18 & 21.64 & $-$6.25 \\
\midrule
\rowcolor{blue!10}
\OURS\mbox{-}1B                        & 48.42 & 64.40 & $-$7.64 \\
\rowcolor{blue!10}
\OURS\mbox{-}3B                        & 49.85 & 64.40 & $-$7.73 \\
\rowcolor{blue!10}
\OURS\mbox{-}8B                        & \textbf{50.28} & \textbf{67.07} & \textbf{$-$7.76} \\
\bottomrule
\end{tabular}
\end{table*}

The unified scientific grammar is the core design of our framework, as shown in Figure~\ref{fig:main}c, built upon the key hypothesis that by jointly modeling multimodal scientific data under a shared vocabulary and unified grammar structure, entities from different scientific modalities can establish explicit semantic associations within a common token space, thereby providing richer knowledge support for downstream tasks. To validate this hypothesis, we conduct experiments from two perspectives: the intrinsic consistency of the grammar and the contribution of grammar components to downstream tasks.

\textbf{Pocket translation task.} As described in Section 2.1.6, we introduce the amino acid–small molecule transformation sequence (the third representation form) in the grammar design of protein pocket data, which explicitly aligns the amino acid-level pocket annotation sequence with the small-molecule-expanded pocket sequence through the directional token \texttt{<Trans>}. To examine whether the model has truly learned the correspondence between the protein symbol system and the small-molecule chemical structure system, we design the pocket translation accuracy as an evaluation metric: given the amino acid-level pocket annotation sequence before the \texttt{<Trans>} token, we evaluate whether the model can correctly "translate" the amino acid residue at each pocket site into its corresponding small-molecule SMILES representation. We report both the site accuracy and sample accuracy which are detailed in~\ref{appendix:translation}. This task requires the model not only to understand the positional semantics of pocket residues within the protein sequence context, but also to master the precise mapping between amino acid identifiers and their molecular structures, thus effectively measuring the effectiveness of the unified grammar in cross-representation knowledge alignment. The evaluation results demonstrate that the model achieves high accuracy on the pocket translation task, confirming that the unified scientific grammar design indeed enables the model to learn systematic correspondences between the protein symbol system and the small-molecule chemical structure system during autoregressive pre-training. Notably, this cross-representation alignment capability is not achieved through additional alignment modules or auxiliary losses, but emerges from the sequence generation learning process under the unified grammar framework, reflecting the intrinsic advantage of the scientific grammar design in facilitating cross-modal knowledge integration.

\textbf{Grammar component ablation.} To understand how different components of the unified grammar synergistically contribute to downstream tasks, we design a progressive comparative experiment on the task of pocket-conditioned ligand generation. As shown in Table~\ref{tab:ablation_pretrain}, we specifically examine three pre-training data configurations:
(i) completely excluding protein ligand-binding sites and protein–ligand complex data; (ii) including these two types of data but removing the pocket translation (\texttt{<Trans>}) data; and (iii) including all grammar components. The performance across these three configurations on the interaction-aware ligand design for binding pockets task exhibits a clear progressive improvement. Under configuration (i), model performance is near random level, indicating that the mere coexistence of different modality data is insufficient for the model to spontaneously establish functional cross-modal associations—even when the pre-training corpus simultaneously contains protein and small molecule data, the model cannot establish the conditional generation mapping between pocket context and ligands without data that explicitly encodes interface interaction relationships. Configuration (ii), which introduces protein ligand-binding sites and protein–ligand complex data, leads to significant performance improvement, demonstrating that these two types of data provide the model with direct supervision for binding site localization and pocket–ligand co-occurrence patterns. However, at this stage the semantic correspondence between pocket residues in the amino acid symbol system and the small-molecule structural system still lacks explicit alignment learning, limiting the model's deep understanding of pocket chemical features. Finally, configuration (iii), which adds pocket translation data, achieves a substantial performance leap. Notably, the pocket translation data fail to involve any ligand information and is solely designed to train the model to translate pocket amino acids into their corresponding SMILES structures, yet it yields the most critical performance gain for ligand generation. This result indicates that after acquiring this alignment knowledge, the model's understanding of pocket residue chemical properties in the ligand generation task no longer relies on isolated pattern matching, but is instead built upon cross-representation semantic equivalence relations, thereby more accurately capturing the complementary constraints between pockets and ligands. This progressive experimental result provides key empirical support for the design philosophy of the unified scientific grammar: the unified grammar is not merely a formal representational unification, but rather achieves genuine knowledge interoperability and synergy across modalities by constructing deep semantic bridges in a shared grammar space. When pre-training directly learns "how to construct inter-object relationships" in a grammar space consistent with downstream tasks, the objective gap between pre-training and downstream applications is effectively eliminated.

Combining the above two experiments, we conclude that the unified scientific grammar not only achieves representational unification of multimodal scientific entities at the formal level, but also facilitates the integration and transfer of deep knowledge across different symbol systems at the semantic level. In particular, through nested grammar structures (pocket tokens embedded within protein grammar, pocket content expanded into small-molecule representations) and explicit cross-representation transformation sequences, the model is able to simultaneously capture protein sequence context information and small-molecule chemical structural features within a unified token space, thereby establishing a solid knowledge foundation for downstream tasks involving cross-modal interactions.

\subsubsection{Cross-Domain Synergy in Multi-Task Fine-Tuning}
\label{sec:cross-domain-ft}

Another core hypothesis of our framework is that under the unified scientific grammar, multi-domain joint modeling enables different domains of scientific data to share underlying structural regularities and compositional logic, thereby producing cross-domain knowledge transfer and mutual gains during joint learning. To validate this hypothesis, we design comparative experiments between cross-domain joint training and single-task independent training during the supervised fine-tuning (SFT) stage.

% \begin{wrapfigure}[25]{r}{0.50\textwidth}
%     \centering
%     \includegraphics[width=\linewidth]{pics/delta_metrics.pdf}
%     \caption{Relative performance gain of multi-task joint SFT over single-task independent SFT across four representative downstream tasks: interaction-aware ligand design for binding pockets (Ligand), protein ligand-binding site identification (Pocket), retrosynthesis prediction (Retrosynthesis), and unconditional material generation (Material). Each bar represents a task-specific evaluation metric. All metrics show consistent positive gains under joint training, indicating that the unified scientific grammar enables effective cross-domain knowledge transfer.}
%     \label{fig:delta_sft_all}
% \end{wrapfigure}

Specifically, we select four representative downstream tasks covering different scientific domains and task types: retrosynthesis prediction, pocket-conditioned ligand generation, binding site identification given a protein sequence, and unconditional material generation. In the joint training setting, we mix the training data of all four tasks and perform unified SFT; in the independent training setting, each task is fine-tuned separately using only its corresponding single-task data.

\begin{wrapfigure}[25]{r}{0.50\textwidth}
    \centering
    \includegraphics[width=\linewidth]{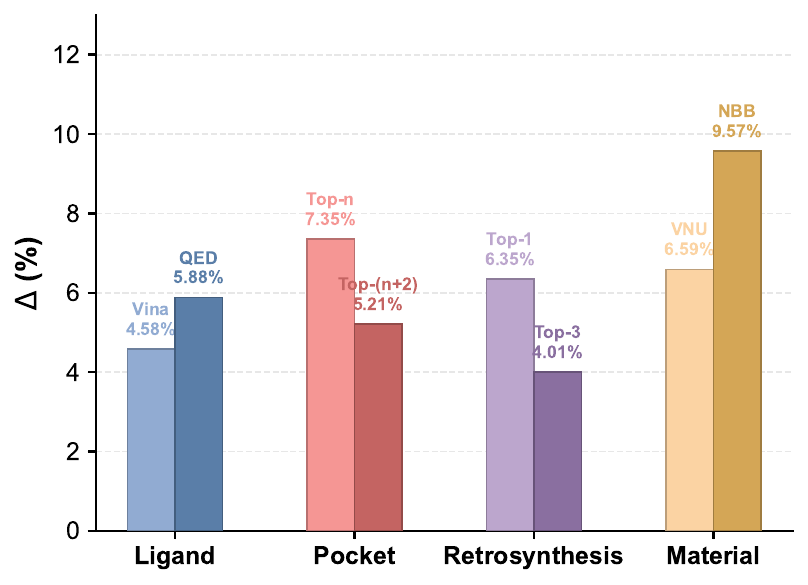}
    \caption{Relative performance gain of multi-task joint SFT over single-task independent SFT across four representative downstream tasks: interaction-aware ligand design for binding pockets (Ligand), protein ligand-binding site identification (Pocket), retrosynthesis prediction (Retrosynthesis), and unconditional material generation (Material). Each bar represents a task-specific evaluation metric. All metrics show consistent positive gains under joint training, indicating that the unified scientific grammar enables effective cross-domain knowledge transfer.}
    \label{fig:delta_sft_all}
\end{wrapfigure}

As shown in Figure~\ref{fig:delta_sft_all}, multi-task joint SFT outperforms the corresponding single-task independent SFT on all four tasks. This finding carries multiple implications. First, from the perspective of knowledge transfer, there indeed exist shareable underlying regularities across tasks from different domains. For example, the molecular transformation and bond-breaking/reorganization patterns learned in retrosynthesis prediction may provide complementary chemical knowledge for molecular scaffold construction in ligand generation; the learning of protein sequence–structure–function relationships in the pocket identification task can directly benefit the understanding of pocket structural constraints in pocket-conditioned ligand generation. 
Second, from a regularization perspective, multi-task joint training naturally serves as implicit regularization by introducing data distributions from different domains, helping to alleviate overfitting that may arise in single-task SFT and enabling the model to learn more generalizable scientific representations. Finally, this result further corroborates the effectiveness of the unified scientific grammar from the downstream fine-tuning stage—precisely because data from different domains are mapped into a shared token space under the unified grammar, the model can effectively capture cross-domain structural commonalities and complementary information during joint learning, rather than treating different tasks as isolated optimization problems. The unified grammar is not merely a formal representational unification, but provides an effective channel for deep knowledge transfer and complementary learning across different scientific domains through shared grammar structures and vocabulary space. This finding provides direct empirical support for the feasibility of the "one model fits all" paradigm in the natural sciences.

\subsection{Model and Training}

\subsubsection{Model Architecture}

Our framework adopts the autoregressive language model as the unified modeling architecture. For the main experiments, we select Qwen3-8B~\citep{yang2025qwen3} to perform joint modeling over multi-domain scientific data. In addition, to explore the effect of model scale on performance (scaling law) and to verify the robustness and generalizability of our framework across different model architectures, we also employ Llama3.2-1B and Llama3.2-3B~\citep{grattafiori2024llama} as supplementary experimental models, forming a model series spanning three parameter scales: 1B, 3B, and 8B.

For vocabulary construction, each model retains the complete vocabulary of its original LLM and extends it with the scientific grammar boundary tokens (special tokens) defined by our framework. For parameter initialization, the parameters of the model backbone and embedding layer inherit the pre-trained weights from the corresponding LLM, while the additional embedding parameters introduced by the newly added special tokens are randomly initialized. As validated in Section 2.2.1, this initialization strategy effectively inherits the general sequence modeling priors of LLMs without incurring additional natural language training overhead, providing beneficial inductive biases for subsequent scientific modality learning.

\subsubsection{Training Pipeline}

As shown in Figure~\ref{fig:main}d, the model training is divided into two stages: continued pre-training and post-training, corresponding to the unified acquisition of multi-domain scientific knowledge and the unified activation of cross-domain generative capabilities, respectively.

\textbf{Continued Pre-training.} In the continued pre-training stage, we use the seven-modality pre-training corpus constructed in Section 2.1 as training data and perform autoregressive continued pre-training on models at all three scales. The optimization objective at this stage is standard next-token prediction, i.e., maximizing the log-likelihood of training sequences. By jointly training on multi-domain data including proteins, antibodies, small molecules, chemical reactions, materials, protein pockets, and protein pocket–ligand complexes under the unified scientific grammar, the model learns intra-modal sequence regularities as well as cross-modal structural associations and semantic correspondences within the shared token sequence space.

\textbf{Post-training.} In the post-training stage, we employ supervised fine-tuning (SFT) to simultaneously activate the model's conditional generation capabilities across multiple representative scientific tasks using a small amount of data within the unified framework. Specifically, we select small training samples from the training set of the evaluation datasets of four representative downstream tasks corresponding to the four representative data forms involved in the pre-training stage—reactions, materials, protein ligand-binding sites, and protein–ligand complexes—namely retrosynthesis prediction, unconditional material generation, protein ligand-binding site identification, and interaction-aware ligand design for binding pockets, and process them into scientific grammar sequence formats consistent with those used during pre-training. On this basis, each training sequence is partitioned according to an "instruction" and "output" template: the portion of the sequence that constitutes the task condition serves as the instruction, while the target portion to be generated serves as the output, with training loss computed only on the token predictions of the output portion. The training data from all four tasks are mixed for unified SFT. As validated in Section~\ref{sec:cross-domain-ft}, this cross-domain joint fine-tuning strategy enables synergistic gains across different tasks, with each task achieving performance superior to single-task independent fine-tuning.
\section{Evaluation}

To systematically evaluate the performance of the LOGOS framework across multi-domain scientific tasks, as shown in Figure~\ref{fig:main}e, we select six representative downstream tasks covering different scientific domains and task types. Based on the correspondence between each task and the sequence formats used during pre-training, these tasks can be divided into two groups. The first group consists of four generative tasks whose formats are directly unified with the pre-training data: interaction-aware ligand design for binding pockets (Section~\ref{sec:sbdd}), protein ligand-binding site identification (Section~\ref{sec:pocket-det}), retrosynthesis prediction (Section~\ref{sec:retro}), and unconditional material generation (Section~\ref{sec:mat}). The input–output formats of these four tasks are fully consistent with the scientific grammar of their corresponding modality data during pre-training, and are designed to validate the effectiveness of the design principle that "the pre-training objective and downstream tasks are formally aligned within a unified grammar space." The second group includes two tasks whose sequence formats are not directly covered during pre-training: protein editing (Section~\ref{sec:prot_edit}) and antibody CDR region design (Section~\ref{sec:cdr}), which are intended to explore the model's ability to generalize to new task formats under the unified scientific grammar framework.

\subsection{Interaction-Aware Ligand Design for Binding Pockets}
\label{sec:sbdd}

\begin{table*}[tb]
\centering
\caption{Results of ligand design for binding pockets. Noted that for
drug candidates, optimal LogP values are typically in the range of -0.4 to 5.6, showing no preference toward either higher or lower values within this interval~\citep{lin2025cbgbench}. }
\label{tab:sbdd_results}
\normalsize
\begin{tabular}{@{}lccccc@{}}
\toprule
\textbf{Method} & \textbf{Vina}$\downarrow$ & \textbf{QED}$\uparrow$ & \textbf{SAS}$\uparrow$ & \textbf{LogP} & \textbf{LPSK}$\uparrow$ \\
\midrule
3DSBDD                                & $-$6.32 & 0.49 & 0.62 & 0.51 & 4.70 \\
GraphBP                               & $-$4.61 & 0.44 & 0.60 & 3.22 & \textbf{4.73} \\
DiffBP                                & $-$7.28 & 0.45 & 0.59 & 5.01 & 4.51 \\
D3FG                                  & $-$6.70 & 0.49 & 0.61 & 1.49 & 4.72 \\
DecompDiff                            & $-$7.13 & 0.50 & 0.63 & 1.25 & 4.41 \\
Pocket2Mol                            & $-$6.95 & 0.50 & 0.69 & 2.39 & 4.55 \\
TargetDiff                            & $-$7.38 & 0.53 & 0.65 & 1.63 & 4.57 \\
TamGen                               & $-$6.79 & 0.55 & 0.70 & 2.55 & 4.68 \\
NatureLM (8$\times$7B)                & $-$6.91 & 0.53 & 0.72 & 0.96 & 4.70 \\
\midrule
\rowcolor{blue!10}
\OURS\mbox{-}1B                        & $-$7.64 & 0.55 & 0.73 & 1.28 & 4.69 \\
\rowcolor{blue!10}
\OURS\mbox{-}3B                        & $-$7.73 & 0.55 & \textbf{0.74} & 1.41 & 4.71 \\
\rowcolor{blue!10}
\rowcolor{blue!10}
\OURS\mbox{-}8B                        & \textbf{$-$7.76} & \textbf{0.57} & 0.73 & 1.22 & 4.71 \\
\bottomrule
\end{tabular}
\end{table*}

Pocket-conditioned ligand generation is one of the most representative tasks in structure-based drug design, aiming to generate small-molecule ligands capable of specifically binding to a protein binding pocket based on its local structural information. In our framework, the input and output formats of this task are fully consistent with the grammar of protein–ligand complex data during pre-training: the input is a protein sequence with pocket residues expanded into SMILES representations \texttt{<ProteinS>...<PocketS>SMILES<PocketE>...<ProteinE><MoleculeS>}, and the output is the ligand SMILES sequence \texttt{SMILES<MoleculeE>}. This design ensures consistency between the pre-training objective and the downstream ligand generation task in both form and optimization direction.

\textbf{Setting.} We conduct experiments on the PDBBind (v2016) dataset~\citep{wang2005pdbbind}. The refined set (excluding the core set)     serves as the training data for this task during the post-training stage, and the core set is used as the test set for evaluation. Evaluation metrics include Vina docking score (Vina$\downarrow$, binding affinity), QED (drug-likeness), SAS (synthetic accessibility), LogP (lipophilicity, with a reasonable range of $-0.4$ to $5.6$~\citet{lin2025cbgbench}), and LPSK (Lipinski's Rule of Five,~\citet{lipinski2012experimental}).

\textbf{Baselines.} Our baselines are divided into two categories. The first category consists of domain-specific methods, including 3DSBDD~\citep{luo20213d}, GraphBP~\citep{liu2022generating}, DiffBP~\citep{lin2025diffbp}, D3FG~\citep{lin2023functional}, DecompDiff~\citep{guan2023decompdiff}, Pocket2Mol~\citep{peng2022pocket2mol}, TargetDiff~\citep{guan20233d}, and TamGen~\citep{hu2025target}. These methods all explicitly rely on three-dimensional atomic coordinates of both the protein and ligand to model pocket–ligand spatial interactions. The second category is NatureLM (8$\times$7B,~\citet{xia2025nature}), which serves as a representative of general-purpose scientific generative models that use natural language as a cross-modal interface.

\begin{figure}[th]
    \centering
    \includegraphics[width=\textwidth]{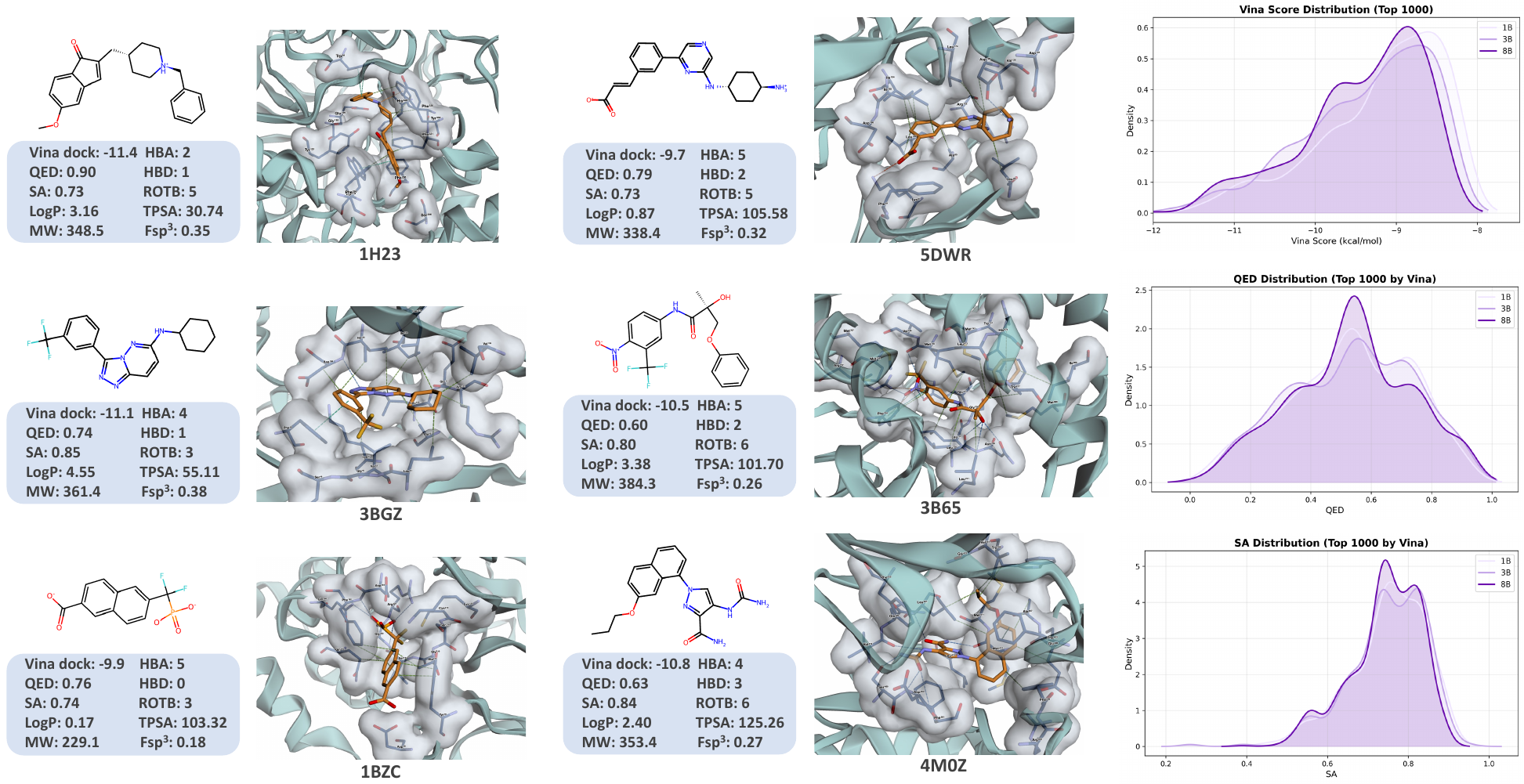}
    \caption{Case study and property analysis of LOGOS-generated pocket-conditioned ligands. \textbf{Left.} Generated ligands for six protein pockets from the PDBBind core set. For each target, the 2D molecular graph, 3D binding pose, and key physicochemical properties are shown. Dashed lines in the 3D views denote non-covalent interactions between the ligand and pocket residues, computed using PLIP~\citep{salentin2015plip}, and color-coded by type: hydrogen bonds (orange), hydrophobic contacts (green), salt bridges (yellow), $\pi$–$\pi$ stacking (cyan), and $\pi$–cation interactions (blue). The generated ligands achieve strong binding affinities with diverse interaction patterns spanning both polar and non-polar contacts, without relying on explicit 3D coordinate inputs. \textbf{Right.} Distributions of Vina score, QED, and SA for the top-1000 generated ligands ranked by Vina docking score, compared across three model scales (1B, 3B, 8B). As model capacity increases, the Vina distribution shifts toward stronger binding affinity, while QED and SA distributions remain comparably favorable, indicating that LOGOS can effectively improve binding strength with scaling without sacrificing drug-likeness or synthetic accessibility.}
    \label{fig:interaction}
\end{figure}

\textbf{Results and Analysis.} As shown in Table~\ref{tab:sbdd_results}, LOGOS achieves superior performance across both binding affinity and multi-dimensional drug-chemical properties.

\textit{Superior binding affinity without explicit geometric dependencies.} On the Vina score, LOGOS-8B achieves the best result of $-7.76$, and LOGOS-1B ($-7.64$) also outperforms all baseline models. Notably, unlike domain-specific methods that explicitly model three-dimensional atomic coordinates with dedicated geometric networks, LOGOS operates in a purely sequential paradigm using discretized token encoding of pocket residue features. This result supports our hypothesis: through grammaticalization and tokenization of spatial interaction relationships, LOGOS can effectively capture pocket–ligand binding patterns without explicit 3D coordinate inputs. Figure~\ref{fig:interaction} (left) further illustrates this capability: the generated ligands form diverse non-covalent interactions across different targets, confirming that LOGOS captures physically meaningful binding patterns in a purely sequential paradigm.

\textit{Parameter efficiency.} Compared with the large language model-based NatureLM (8$\times$7B, Vina $-6.91$), LOGOS-1B achieves a notably better Vina score of $-7.64$ with approximately 1/56 of the total parameters. This echoes the capacity allocation analysis in Section 2.2.1: by adopting scientific grammar as the cross-modal interface, LOGOS concentrates model capacity on native scientific representation learning, yielding higher parameter efficiency for domain-native tasks.

\textit{Advantages across multi-dimensional drug-chemical properties.} Beyond binding affinity, LOGOS also performs favorably in drug-likeness (QED: 0.57 for LOGOS-8B) and synthetic accessibility (SAS: 0.74 for LOGOS-3B). As shown in Figure~\ref{fig:interaction} (left), the generated ligands consistently exhibit favorable physicochemical profiles across diverse pockets, satisfying multiple drug-likeness criteria simultaneously. We attribute this in part to the synergistic effect of multi-domain pre-training: small molecule data provides structural regularities of valid chemical space; reaction data offers complementary bond-reorganization knowledge for scaffold construction; and pocket translation data strengthens fine-grained understanding of residue chemical features.

\textit{Scaling behavior.} From 1B to 8B, both binding affinity and drug-chemical properties show consistent improvement with increasing model capacity. As illustrated in Figure~\ref{fig:interaction} (right), the Vina score distribution progressively shifts toward stronger binding affinity, while QED and SA distributions remain comparably favorable, confirming that LOGOS scales binding strength without sacrificing drug-likeness or synthetic accessibility.

\subsection{Protein Ligand-Binding Site Identification}
\label{sec:pocket-det}

\begin{table*}[tb]
\centering
\caption{Results of protein ligand-binding site identification on COACH420 and HOLO4K. While all baselines require 3D protein structures as input, LOGOS operates solely on amino acid sequences, yet LOGOS-8B achieves performance second only to P2Rank, whose predictions serve as the training annotations of LOGOS.}
\label{tab:pocket}
\small
\begin{tabular}{@{}lcccc@{}}
\toprule
& \multicolumn{2}{c}{\textbf{COACH420}} & \multicolumn{2}{c}{\textbf{HOLO4K}} \\
\cmidrule(lr){2-3} \cmidrule(lr){4-5}
\textbf{Method} & \textbf{Top-$n$}$\uparrow$ & \textbf{Top-($n$+2)}$\uparrow$ & \textbf{Top-$n$}$\uparrow$ & \textbf{Top-($n$+2)}$\uparrow$ \\
\midrule
Fpocket                               & 56.4 & 68.9 & 52.4 & 63.1 \\
SiteHound                             & 53.0 & 69.3 & 50.1 & 62.1 \\
MetaPocket 2.0                        & 63.4 & 74.6 & 57.9 & 68.6 \\
DeepSite                              & 56.4 & 63.4 & 45.6 & 48.2 \\
\rowcolor{gray!10}
P2Rank                                & 72.0 & 78.3 & 68.6 & 74.0 \\
\midrule
\rowcolor{blue!10}
\OURS\mbox{-}1B                       & 59.2 & 66.5 & 58.0 & 67.7 \\
\rowcolor{blue!10}
\OURS\mbox{-}3B                       & 65.0 & 70.6 & 58.2 & 68.0 \\
\rowcolor{blue!10}
\OURS\mbox{-}8B                       & \textbf{66.5} & \textbf{74.8} & \textbf{58.5} & \textbf{68.9} \\
\bottomrule
\end{tabular}
\end{table*}

Protein ligand-binding site identification aims to predict potential small-molecule binding pockets from a protein, serving as a critical prerequisite step for virtual screening and lead compound discovery in structure-based drug design. In our framework, this task directly reuses the representation form (binding site identification sequence) from the protein pocket data during pre-training: the input is a complete protein sequence without pocket annotations followed by a task semantic token \texttt{<ProteinS>...<ProteinE><Search>}, and the output is a protein sequence containing pocket position annotations \texttt{<ProteinS>...<PocketS>...<PocketE>...<ProteinE>}. That is, the model generates pocket boundary tokens on the protein sequence in an autoregressive manner, thereby transforming the pocket identification problem into a sequence generation task consistent with the pre-training objective.

\textbf{Setting.} We follow the evaluation protocol of P2Rank~\citep{krivak2018p2rank} and conduct evaluation on two benchmark datasets: COACH420~\citep{yang2013protein} and HOLO4K~\citep{schmidtke2010large}. To balance the proportion of data from different domains in the post-training stage, the training data for this task is constructed by merging the CHEN11 dataset~\citep{chen2011critical} with a sampled subset of the data used in the continued pre-training stage. The evaluation metric follows the DCC criterion (distance from the predicted pocket center to the nearest ligand atom, with a threshold of 4~\AA), and we report identification success rates under two settings: Top-$n$ and Top-($n$+2), where $n$ is the actual number of ligands in the protein structure. Top-$n$ only examines the model's top-$n$ ranked predicted pockets, requiring the prediction precision to strictly match the actual number of binding sites; Top-($n$+2) allows 2 additional redundant predictions, measuring the model's recall capability under moderately relaxed conditions~\citep{krivak2018p2rank}.

\textbf{Baselines.} The comparison methods include Fpocket~\citep{le2009fpocket}, SiteHound~\citep{hernandez2009sitehound}, MetaPocket 2.0~\citep{zhang2011identification}, DeepSite~\citep{jimenez2017deepsite}, and P2Rank~\citep{krivak2018p2rank}. All of these methods take the three-dimensional structure of the protein as input and rely on explicit spatial geometric information for pocket localization.

\begin{figure}[th]
    \centering
    \includegraphics[width=\textwidth]{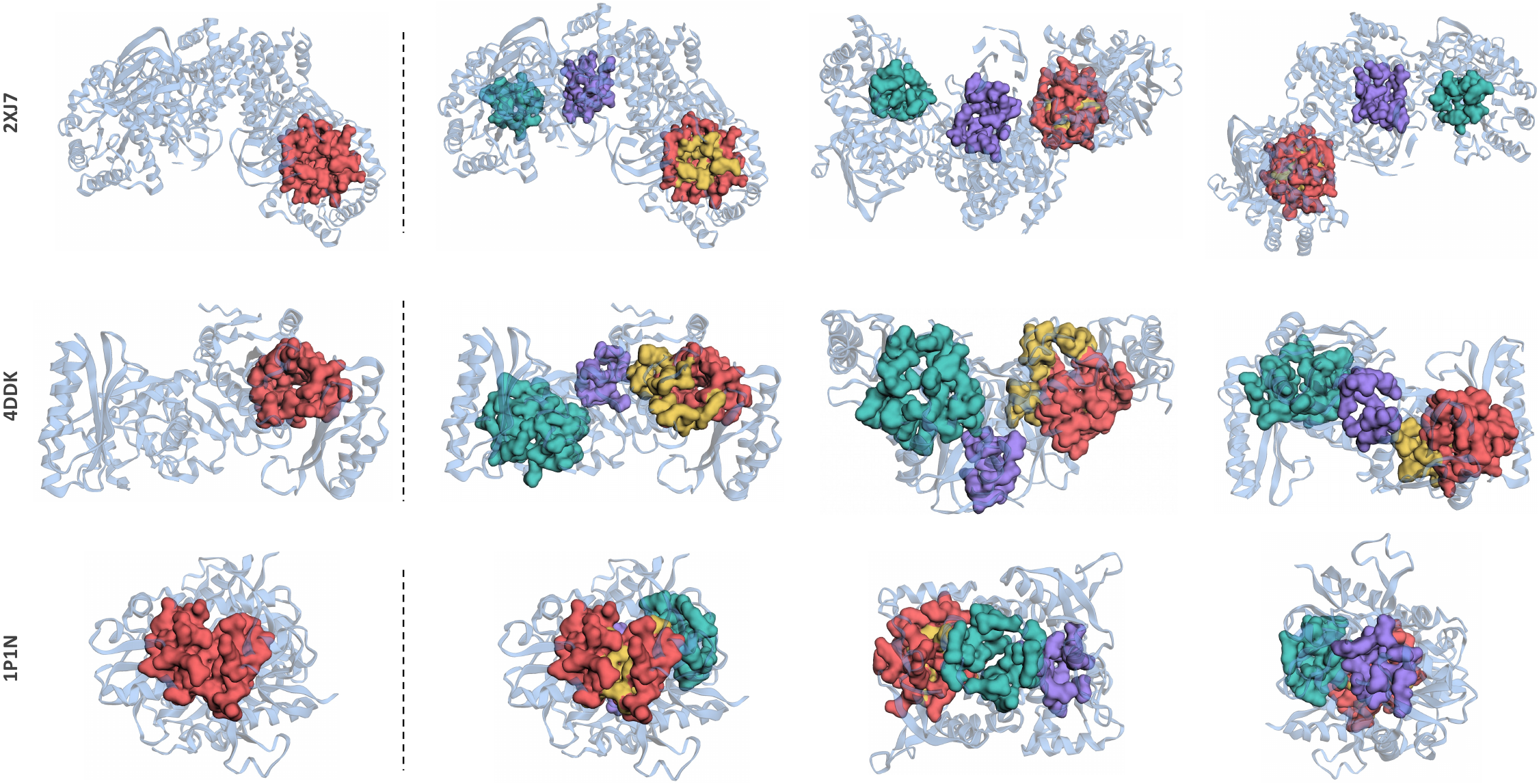}
    \caption{Visualization of protein ligand-binding site identification by LOGOS on three protein cases from the PDBbind core set. Each row corresponds to one protein (PDB ID: 2XJ7, 4DDK, and 1P1N). \textbf{Left.} The ground-truth binding pocket (red) annotated in the dataset. \textbf{Right.} The Top-3 pockets predicted by LOGOS (yellow, purple, and teal), viewed from three different angles to better illustrate the spatial distribution of identified sites. Notably, LOGOS not only recovers the annotated ground-truth pocket but also identifies additional putative binding sites on the protein surface using only the amino acid sequence as input.}
    \label{fig:pocket}
\end{figure}

\textbf{Results and Analysis.} As shown in Table~\ref{tab:pocket}, LOGOS achieves performance second only to P2Rank on both datasets, outperforming all other comparison methods.

\textit{Competitive performance under a purely sequential paradigm.} On both COACH420 and HOLO4K, LOGOS-8B surpasses all baseline methods except P2Rank on Top-$n$ and Top-($n$+2) metrics, achieving results second only to P2Rank. It is particularly noteworthy that all comparison methods require the three-dimensional atomic coordinates of the protein as a necessary input, whereas LOGOS achieves competitive pocket identification performance using only the one-dimensional amino acid sequence as input. As shown in Figure~\ref{fig:pocket}, LOGOS recovers the ground-truth pocket and identifies additional potential binding sites on the protein surface using only the amino acid sequence in each case. This means that LOGOS extends the applicability of binding site identification from "proteins with resolved 3D structures" to "proteins with only sequence information available"—considering that the number of known protein sequences far exceeds the number of resolved structures, this capability carries significant practical value.

\textit{Analysis of the performance gap with P2Rank.} A performance gap remains between LOGOS and P2Rank. As described in Section 2.1.6, the pocket annotations used during our pre-training stage are derived from P2Rank predictions on PDB structures, which naturally upper-bounds the achievable performance. Nevertheless, the two approaches differ fundamentally at the methodological level: P2Rank requires complete three-dimensional structural information at inference time, whereas LOGOS operates solely on amino acid sequences, enabling pocket identification for the much larger space of proteins without resolved structures. It effectively shifts pocket identification from a 3D structure-dependent paradigm to a purely sequence-based one, greatly expanding the applicability of this task.

\textit{Scaling behavior and cross-domain knowledge gains.} On both datasets, LOGOS exhibits a consistent performance improvement trend from 1B to 8B. Furthermore, as an interface task between protein sequences and the small-molecule chemical space, pocket identification performance also benefits from multi-domain joint learning: pocket–ligand complex data during pre-training provides functional context for pockets, while pocket translation data strengthens the understanding of pocket residue chemical features. These two complementary sources of knowledge are expected to jointly facilitate the inference of binding sites.

\subsection{Retrosynthesis Prediction}
\label{sec:retro}

\begin{table*}[tb]
\centering
\caption{Results of retrosynthesis prediction.}
\label{tab:retro_results}
\small
\begin{tabular}{@{}lcc@{}}
\toprule
\textbf{Method} & \textbf{Top-1}$\uparrow$ & \textbf{Top-3}$\uparrow$ \\
\midrule
LocalRetro                            & 51.5\% & 76.5\% \\
R\mbox{-}SMILES                       & 56.0\% & 79.1\% \\
EditRetro                             & 60.8\% & 80.6\% \\
NatureLM (1B)                         & 68.6\% & 86.8\% \\
NatureLM (8B)                         & 70.2\% & 85.9\% \\
NatureLM (8$\times$7B)                & 71.9\% & 87.4\% \\
\midrule
\rowcolor{blue!10}
\OURS\mbox{-}1B                       & 64.0\% & 72.4\% \\
\rowcolor{blue!10}
\OURS\mbox{-}3B                       & 71.5\% & 74.2\% \\
\rowcolor{blue!10}
\OURS\mbox{-}8B                       & \textbf{74.8\%} & 75.6\% \\
\bottomrule
\end{tabular}
\end{table*}

Retrosynthesis prediction aims to infer possible reactant combinations given a target product molecule, and is a core task in synthetic route planning and medicinal chemistry. In our framework, this task directly reuses the reverse grammar form of the chemical reaction data from the pre-training stage: the input is the SMILES sequence of the product molecule followed by a task token \texttt{<MoleculeS>SMILES<MoleculeE>  <ReverseReact>}, and the output is the SMILES sequence of the reactants \texttt{<MoleculeS>SMILES<MoleculeE>}. This format is fully consistent with the reaction data during pre-training, ensuring unification in both grammar structure and optimization objective.

\textbf{Setting.} We conduct experiments on USPTO-50K~\citep{lowe2012extraction, schneider2016s}. Its training set is used as the training data for this task during unified post-training, and evaluation is performed on the test set. We report Top-1 and Top-3 accuracy as evaluation metrics.

\textbf{Baselines.} The comparison methods are divided into two categories. The first category consists of domain-specific methods, including LocalRetro~\citep{chen2021deep}, R-SMILES~\citep{zhong2022root}, and EditRetro~\citep{han2024retrosynthesis}. The second category is NatureLM (1B / 8B / 8$\times$7B), serving as a representative of general-purpose scientific models that use natural language as a cross-modal interface.

\begin{figure}[th]
    \centering
    \includegraphics[width=\textwidth]{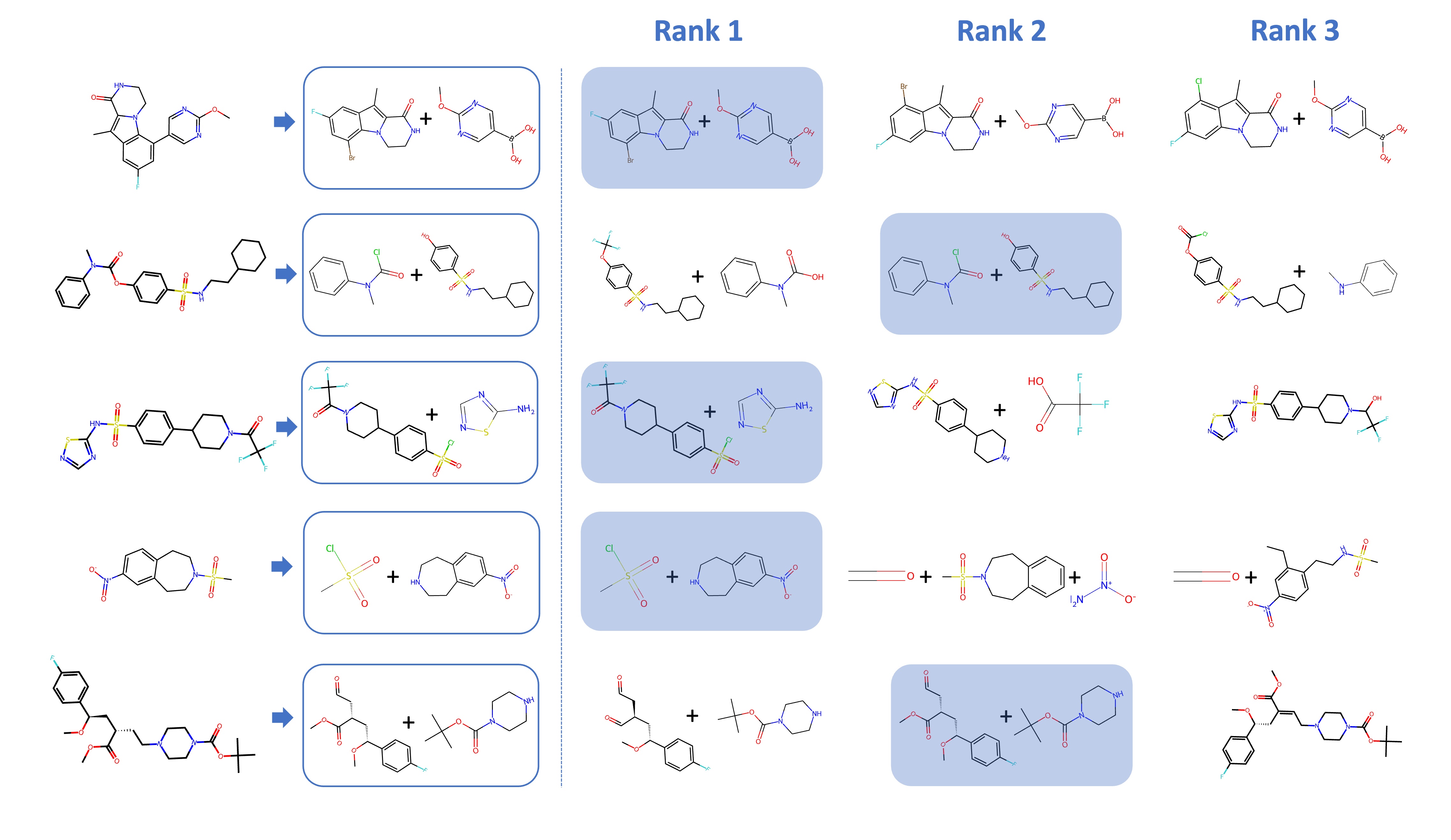}
    \caption{Case study of retrosynthesis predictions by LOGOS on USPTO-50K test set. \textbf{Left.} The target product molecule and its ground-truth reactants. \textbf{Right.} The top-3 predictions generated by LOGOS. Blue-shaded boxes indicate predictions that match the ground truth.}
    \label{fig:reaction}
\end{figure}

\textbf{Results and Analysis.} As shown in Table~\ref{tab:retro_results}, LOGOS-8B achieves the best Top-1 accuracy of 74.8\% among all compared methods.

\textit{Comprehensive lead in Top-1 accuracy.} Top-1 accuracy directly reflects a model's ability to identify the most probable retrosynthetic pathway, and is often the most relevant metric in practical synthesis planning. LOGOS-8B (74.8\%) outperforms all compared methods in Top-1 accuracy, including NatureLM (8$\times$7B) (71.9\%) and representative domain-specific approaches. As shown in Figure~\ref{fig:reaction}, LOGOS correctly recovers the ground-truth reactants as its top-ranked prediction in three out of five sampled cases. It suggests that modeling directly in the native representation space of chemical transformations may allow the model to concentrate its probability mass on the most plausible bond-breaking and reorganization patterns, potentially contributing to higher confidence in top-ranked predictions. We note that the Top-3 accuracy of LOGOS does not reach the same level of advantage, which may partly reflect a trade-off between prediction precision and candidate diversity inherent to different generation paradigms.

\textit{Parameter efficiency and scaling behavior.} LOGOS-3B achieves competitive Top-1 accuracy with a considerably smaller parameter budget, while LOGOS-8B further extends this advantage to 74.8\%. From 1B to 8B, Top-1 accuracy exhibits a continuous and substantial improvement (64.0\% $\rightarrow$ 71.5\% $\rightarrow$ 74.8\%), indicating that LOGOS possesses favorable scaling properties in chemical transformation modeling. This scaling behavior observed in the retrosynthesis task is consistent with other tasks such as ligand design, suggesting that LOGOS can stably benefit from increased parameter scale across tasks in different domains.

\subsection{Unconditional Material Generation}
\label{sec:mat}

\begin{table*}[tb]
\centering
\caption{Results of MOF generation.}
\label{tab:mof_results}
\small
\begin{tabular}{@{}lccc@{}}
\toprule
\textbf{Method} & \textbf{Valid}$\uparrow$ & \textbf{VNU}$\uparrow$ & \textbf{NBB}$\uparrow$ \\
\midrule
MOFDiff                                & 10.13 & 7.95 & 0.00 \\
Mofflow\mbox{-}2                       & 38.84 & 31.35 & 10.10 \\
\midrule
\rowcolor{blue!10}
\OURS\mbox{-}1B                        & 44.46 & 37.96 & 16.77 \\
\rowcolor{blue!10}
\OURS\mbox{-}3B                        & 45.14 & 38.97 & 17.23 \\
\rowcolor{blue!10}
\OURS\mbox{-}8B                        & \textbf{45.19} & \textbf{39.02} & \textbf{17.78} \\
\bottomrule
\end{tabular}
\end{table*}

Unconditional material generation aims to sample and generate metal-organic frameworks (MOFs) that are chemically valid and structurally novel from the learned material structure distribution. In our framework, the input–output format of this task is consistent with the grammar of material data during pre-training: given the prompt \texttt{<MaterialS>}, the model outputs a complete material sequence \texttt{<MetalS>...<MetalE>...<MoleculeS>...<MoleculeE>...<MaterialE>}, where metal clusters are represented with dedicated tokens \texttt{<MetalS>...<MetalE>} and organic linkers are encoded by reusing the small molecule grammar \texttt{<MoleculeS>...<MoleculeE>}.

\textbf{Setting.} To balance the proportion of data from different domains in the post-training stage, the training data for this task is directly inherited from the materials data used during continued pre-training. We follow the evaluation protocol of MOFFlow-2~\citep{kim2026flexible} and adopt three representative metrics for evaluation: Valid (the proportion of generated MOFs that pass the chemical validity check of MOFChecker~\citep{jin2025mofchecker}), VNU (the proportion of generated MOFs that are simultaneously chemically valid, structurally novel—i.e., their MOFid~\citep{bucior2019identification} does not appear in the training set—and structurally unique after deduplication), and NBB (the proportion of valid generated MOFs that contain at least one novel building block not seen in the training set). These three metrics evaluate generation quality at three progressively stricter levels: chemical validity, structural novelty, and component innovation.

\textbf{Baselines.} The comparison methods are all domain-specific MOF generation models, including the diffusion-based MOFDiff~\citep{fu2024mofdiff} and the flow-matching-based MOFFlow-2~\citep{kim2026flexible}.

\begin{figure}[htbp]
    \centering
    \includegraphics[width=\textwidth]{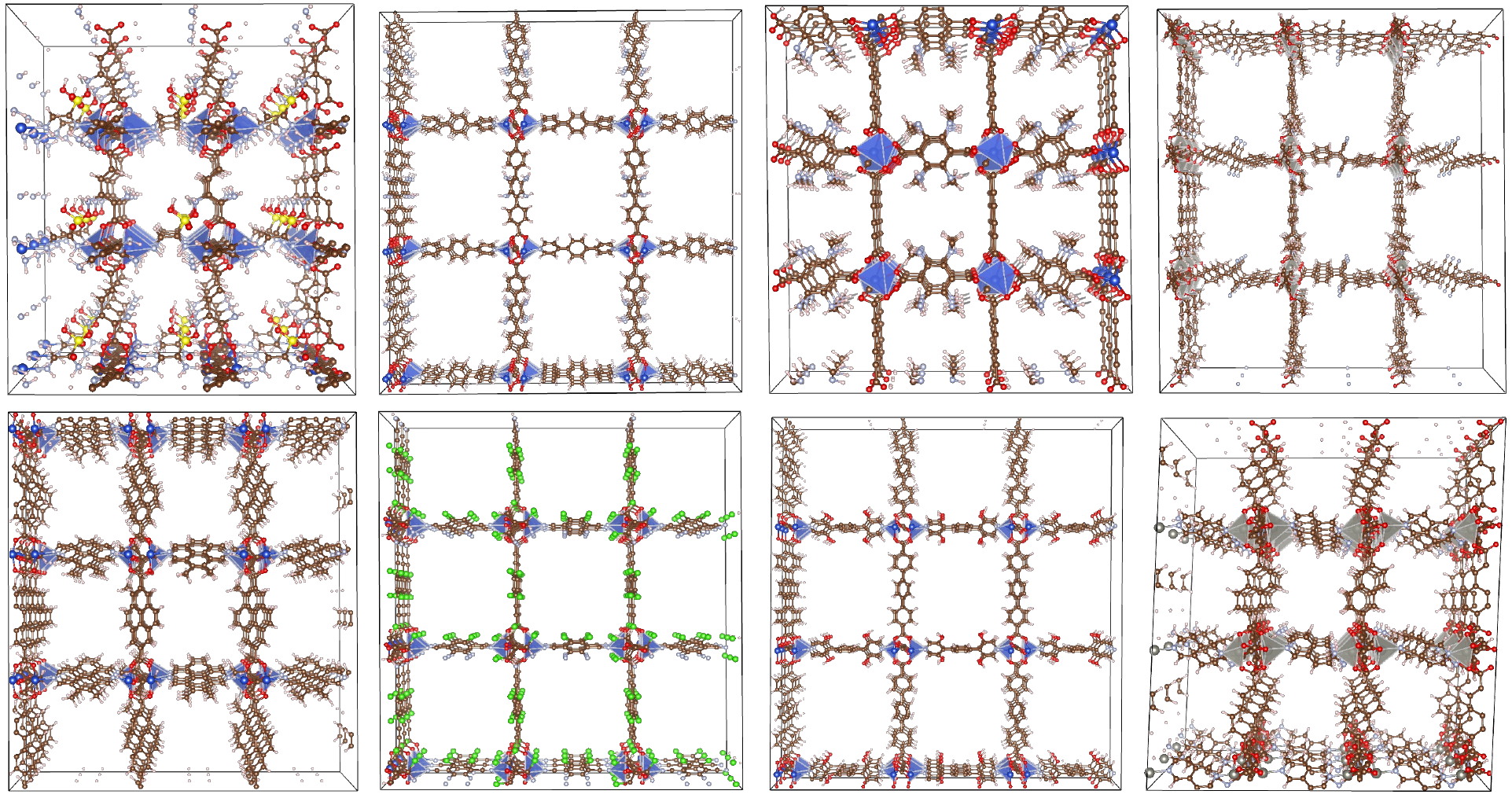}
    \caption{Visualization of MOFs generated by LOGOS. The well-formed crystalline frameworks confirm that the generated sequences faithfully encode valid and diverse MOF compositions. Notably, some structures contain novel building blocks beyond the MOF training set, indicating that the model captures transferable chemical knowledge.}
    \label{fig:material_vis}
\end{figure}

\textbf{Results and Analysis.} As shown in Table~\ref{tab:mof_results}, LOGOS significantly outperforms all domain-specific baseline methods across all three metrics. 

\textit{Breakthrough in component innovation capability.} The improvement in the NBB metric merits particular attention. The application potential of MOFs is closely tied to the expandability of their compositional space: the combinatorial pairing of diverse metal clusters and organic linkers gives rise to a vast candidate material space~\citep{kim2026flexible}, and the discovery of new building blocks is a prerequisite for accessing novel structures and properties. The NBB metric requires that valid generated MOFs contain building blocks absent from the training set, demanding the model not only learn reasonable combinations of known components but also explore previously unseen molecular building blocks. LOGOS-8B achieves an NBB of 17.78\%, a 76\% relative improvement over MOFFlow-2 (10.10\%) and a marked gain over MOFDiff (0.00\%). As shown in Figure~\ref{fig:material_vis}, the generated MOFs exhibit well-formed three-dimensional crystalline frameworks with clearly resolved periodic connectivity, confirming that the output sequences faithfully encode valid and structurally diverse MOF compositions. Notably, several visualized structures contain building blocks absent from the training set, providing direct evidence that the model can generate novel building blocks that are chemically plausible and geometrically coherent. This result supports our central thesis: the impact of artificial intelligence on the natural sciences should extend beyond modeling known systems to generative capabilities oriented toward unexplored chemical spaces.

\textit{Cross-domain synergy under unified scientific grammar.} The material generation task provides an intuitive illustration of cross-domain synergistic effects enabled by the unified scientific grammar. The nested grammar of MOFs—material-level boundary tokens enclosing metal cluster and small molecule tokens—naturally bridges the materials and small molecule domains at the representation level. The SMILES regularities and molecular transformation patterns learned from the small molecule and chemical reaction corpora during pre-training can flow into material generation through the shared \texttt{<MoleculeS>...<MoleculeE>} grammar channel, providing rich chemical priors for constructing reasonable organic linkers. This demonstrates that the unified grammar not only achieves multi-domain representational unification at the formal level, but also enables effective knowledge transfer across domains at the semantic level.

\subsection{Generalization Beyond Pre-training Task Formats}

The preceding four tasks validate the performance of our framework within unified data formats. However, a more challenging question arises: when the sequence format of a downstream task is not explicitly covered during pre-training, can the model leverage the cross-modal knowledge acquired through the unified scientific grammar and generalize to new task formats with only a small amount of supervised fine-tuning? The following two tasks—protein editing and antibody CDR region design—are intended to provide empirical examination of this question.

\subsubsection{Protein Editing}
\label{sec:prot_edit}

\begin{table*}[tb]
\centering
\caption{Results of protein editing on AAV and GFP. Noted that higher diversity and novelty are not equivalent to better performance, but provide insight into the exploration and exploitation trade-offs of different methods~\citep{DBLP:conf/iclr/KirjnerYSBJBF24}.}
\label{tab:protein_fitness}
\small
\begin{tabular}{@{}lcccccc@{}}
\toprule
 & \multicolumn{3}{c}{\textbf{Medium Difficulty}} & \multicolumn{3}{c}{\textbf{Hard Difficulty}} \\
\cmidrule(lr){2-4} \cmidrule(lr){5-7}
\textbf{Method} & \textbf{Fitness}$\uparrow$ & \textbf{Diversity} & \textbf{Novelty} & \textbf{Fitness}$\uparrow$ & \textbf{Diversity} & \textbf{Novelty} \\
\midrule
\multicolumn{7}{c}{\textit{AAV}} \\
\midrule
GFN\mbox{-}AL       & 0.11 & 9.18  & 19.0 & 0.06 & 11.28 & 20.0 \\
CbAS               & 0.24 & 13.05 & 7.5  & 0.20 & 14.82 & 8.3  \\
AdaLead            & 0.26 & 8.23  & 3.1  & 0.22 & 8.15  & 3.0  \\
BOqei              & 0.22 & 15.54 & 0.0  & 0.18 & 17.53 & 0.0  \\
CoMS               & 0.21 & 10.48 & 7.9  & 0.14 & 11.14 & 10.3 \\
PEX                & 0.23 & 2.46  & 1.1  & 0.17 & 3.12  & 1.6  \\
GGS                & 0.29 & 4.37  & 5.0  & 0.33 & 4.19  & 7.4  \\
\rowcolor{blue!10}
\OURS\mbox{-}1B    & 0.69 & 5.30  & 6.5  & 0.69 & 5.20  & 7.8  \\
\rowcolor{blue!10}
\OURS\mbox{-}3B    & 0.69 & 4.50  & 6.5  & 0.70 & 5.00  & 7.82 \\
\rowcolor{blue!10}
\OURS\mbox{-}8B    & \textbf{0.69} & 4.46  & 6.0  & \textbf{0.70} & 4.00  & 8.0  \\
\midrule
\multicolumn{7}{c}{\textit{GFP}} \\
\midrule
GFN\mbox{-}AL       & 0.04 & 24.68 & 213.4 & 0.05 & 23.18 & 213.5 \\
CbAS               & 0.06 & 10.07 & 6.9  & 0.08 & 9.24  & 8.1  \\
AdaLead            & 0.26 & 3.17  & 2.3  & 0.06 & 5.96  & 2.5  \\
BOqei              & 0.09 & 19.72 & 0.0  & 0.00 & 94.17 & 53.8 \\
CoMS               & 0.00 & 132.56 & 191.6 & 0.00 & 144.48 & 201.4 \\
PEX                & 0.22 & 3.38  & 1.0  & 0.00 & 2.62  & 1.0  \\
GGS                & 0.35 & 3.31  & 5.4  & 0.34 & 3.93  & 7.6  \\
\rowcolor{blue!10}
\OURS\mbox{-}1B    & 0.89 & 5.41  & 7.0  & 0.91 & 5.15  & 8.0  \\
\rowcolor{blue!10}
\OURS\mbox{-}3B    & 0.92 & 4.72  & 7.0  & 0.93 & 5.01  & 8.0  \\
\rowcolor{blue!10}
\OURS\mbox{-}8B    & \textbf{0.93} & 4.64  & 7.0  & \textbf{0.93} & 4.76  & 8.0  \\
\bottomrule
\end{tabular}
\end{table*}

Protein editing aims to take a protein sequence with suboptimal functional properties and generate an optimized variant with improved performance on a target attribute, such as fluorescence activity or capsid assembly fitness. Since the sequence format of this task is not directly covered during pre-training, it serves as a rigorous evaluation of the model's capacity to generalize to unseen task formats.

\textbf{Setting.} We follow the experimental setup of GGS~\citep{kirjner2024improving} and select two benchmark datasets—green fluorescent protein (GFP) and adeno-associated virus capsid protein (AAV)—for evaluation, testing under both Medium and Hard difficulty settings. Difficulty is jointly defined by the fitness percentile range of the starting sequence and the minimum number of mutations (Gap) required to reach the 99th fitness percentile: the Medium setting uses starting sequences in the 20th–40th percentile (Gap = 6), and the Hard setting uses starting sequences below the 30th percentile (Gap = 7). Each training sample consists of a starting sequence of inferior fitness paired with a target sequence at the 99th fitness percentile, constituting a directed optimization pair from inferior to superior fitness. Data splitting follows 10-fold cross-validation, where each fold uses a non-overlapping 1/10 as the test set and the remaining 9/10 as training data, with final results averaged over 10 runs. Evaluation metrics include: Fitness (normalized fitness); Diversity (the median edit distance across all sample pairs in the generated sequence set, quantifying intra-set sequence variation); and Novelty (the median of the minimum edit distance from each generated sequence to all starting sequences, quantifying the degree of divergence from the input sequences). In terms of grammar design, we introduce task-specific semantic tokens \texttt{<OptimizeAAV>} and \texttt{<OptimizeGFP>} for the protein editing task to explicitly encode the optimization objective. The input is the sequence to be optimized \texttt{<ProteinS>original sequence<ProteinE><OptimizeAAV/GFP>}, and the output is the optimized sequence \texttt{<ProteinS>optimized sequence<ProteinE>}. This format reuses the boundary tokens of the protein modality, incorporating protein editing into the generative sequence prediction framework solely through the introduction of task-specific semantic tokens.

\textbf{Baselines.} We compare our LOGOS with GFN-AL~\citep{jain2022biological}, CbAS~\citep{brookes2019conditioning}, AdaLead~\citep{sinai2020adalead}, BOqei~\citep{wilson2017reparameterization}, CoMS~\citep{trabucco2021conservative}, PEX~\citep{ren2022proximal}, and GGS~\citep{kirjner2024improving}.

\textbf{Results and Analysis.} As shown in Table~\ref{tab:protein_fitness}, LOGOS consistently achieves the highest Fitness scores across all four settings, surpassing prior methods by a substantial margin. Notably, LOGOS exhibits strong robustness to increased task difficulty: its Fitness remains nearly unchanged between the Medium and Hard settings on both datasets, whereas several baselines incur considerable performance degradation as the optimization gap widens. Beyond Fitness, LOGOS maintains Diversity and Novelty within reasonable ranges. In particular, its Novelty values approach or exceed the Gap thresholds of the corresponding difficulty settings, indicating that the generated sequences introduce meaningful mutational distance rather than constituting trivial perturbations of the input. We note that elevated Diversity and Novelty do not necessarily correlate with functional improvement; certain methods produce highly diverse outputs yet achieve limited Fitness, suggesting that excessive sequence variation without functional guidance may lead to departures from the viable fitness landscape. Overall, LOGOS achieves markedly superior and more stable functional optimization while preserving adequate sequence diversity and novelty.

\textit{Generalization beyond pre-training formats.} The sequence format of protein editing is not explicitly covered during pre-training. Yet LOGOS outperforms specialized methods on this task with a small amount of SFT data, suggesting that sequential regularities and mutation-related knowledge learned from large-scale protein corpora can transfer to unseen task formats. We attribute this to pre-training over the protein modality, which enables the model to internalize compositional constraints and positional dependencies of amino acid sequences; these are then adapted to the editing scenario through task-specific semantic tokens during SFT. This result supports the generalization capacity of the unified scientific grammar framework: when the relevant scientific entities and their modality grammars have been modeled during pre-training, the model can accommodate new task formats through minimal format-level adaptation.

\subsubsection{Antibody CDR Design}
\label{sec:cdr}

\begin{table*}[tb]
\centering
\caption{Results of sequence design on SAbDab dataset.}
\label{tab:sabdab}
\resizebox{\textwidth}{!}{
\Large
\begin{tabular}{@{}lccccccccc@{}}
\toprule
 & \multicolumn{3}{c}{\textbf{CDR-H1}} & \multicolumn{3}{c}{\textbf{CDR-H2}} & \multicolumn{3}{c}{\textbf{CDR-H3}} \\
\cmidrule(lr){2-4} \cmidrule(lr){5-7} \cmidrule(lr){8-10}
\textbf{Method} & AAR(\%)$\uparrow$ & scRMSD$\downarrow$ & plausibility$\uparrow$ & AAR(\%)$\uparrow$ & scRMSD$\downarrow$ & plausibility$\uparrow$ & AAR(\%)$\uparrow$ & scRMSD$\downarrow$ & plausibility$\uparrow$ \\
\midrule
Grafting          & 58.05 & 0.83 & $-$0.597 & 31.46 & 0.79 & $-$0.619 & 19.63 & 3.20 & $-$0.591 \\
ProteinMPNN       & 58.58 & 0.64 & $-$0.603 & 53.18 & 0.61 & $-$0.568 & 41.77 & 2.27 & $-$0.605 \\
ESM\mbox{-}IF1    & 53.80 & 0.66 & $-$0.610 & 46.66 & 0.63 & $-$0.589 & 29.82 & 2.59 & $-$0.607 \\
Diffab\mbox{-}fix & 74.93 & 0.66 & $-$0.512 & 65.41 & 0.59 & $-$0.532 & 49.17 & 2.24 & $-$0.541 \\
AbMPNN            & 72.83 & 1.09 & $-$0.664 & 65.33 & 0.93 & $-$0.677 & 52.99 & 2.80 & $-$0.675 \\
RADAb             & 76.57 & 0.61 & $-$0.505 & 66.16 & 0.57 & $-$0.530 & \textbf{57.02} & \textbf{2.23} & \textbf{$-$0.530} \\
\midrule
\rowcolor{blue!10}
\OURS\mbox{-}1B   & 75.19 & 0.65 & $-$0.508 & 65.31 & 0.59 & $-$0.527 & 42.67 & 2.52 & $-$0.560 \\
\rowcolor{blue!10}
\OURS\mbox{-}3B   & 78.28 & 0.60 & $-$0.496 & 66.14 & 0.57 & $-$0.522 & 44.15 & 2.48 & $-$0.542 \\
\rowcolor{blue!10}
\OURS\mbox{-}8B   & \textbf{79.82} & \textbf{0.57} & \textbf{$-$0.488} & \textbf{66.83} & \textbf{0.55} & \textbf{$-$0.515} & 46.95 & 2.43 & $-$0.533 \\
\midrule
 & \multicolumn{3}{c}{\textbf{CDR-L1}} & \multicolumn{3}{c}{\textbf{CDR-L2}} & \multicolumn{3}{c}{\textbf{CDR-L3}} \\
\cmidrule(lr){2-4} \cmidrule(lr){5-7} \cmidrule(lr){8-10}
\textbf{Method} & AAR(\%)$\uparrow$ & scRMSD$\downarrow$ & plausibility$\uparrow$ & AAR(\%)$\uparrow$ & scRMSD$\downarrow$ & plausibility$\uparrow$ & AAR(\%)$\uparrow$ & scRMSD$\downarrow$ & plausibility$\uparrow$ \\
\midrule
Grafting          & 68.53 & 0.85 & $-$0.506 & 43.19 & 0.52 & $-$0.573 & 43.61 & 1.08 & $-$0.395 \\
ProteinMPNN       & 45.60 & 0.59 & $-$0.612 & 46.78 & 0.46 & $-$0.527 & 47.21 & 0.98 & $-$0.543 \\
ESM\mbox{-}IF1    & 40.97 & 0.61 & $-$0.650 & 43.40 & 0.43 & $-$0.542 & 38.93 & 0.92 & $-$0.569 \\
Diffab\mbox{-}fix & 79.78 & 0.56 & $-$0.386 & 81.19 & 0.44 & $-$0.398 & 67.97 & 0.88 & $-$0.414 \\
AbMPNN            & 75.06 & 0.73 & $-$0.543 & 71.63 & 0.56 & $-$0.528 & 64.51 & 0.91 & $-$0.544 \\
RADAb             & 83.72 & 0.54 & $-$0.379 & 84.58 & 0.44 & $-$0.384 & \textbf{73.11} & \textbf{0.87} & \textbf{$-$0.384} \\
\midrule
\rowcolor{blue!10}
\OURS\mbox{-}1B   & 74.02 & 0.74 & $-$0.536 & 83.16 & 0.44 & $-$0.392 & 51.56 & 0.96 & $-$0.429 \\
\rowcolor{blue!10}
\OURS\mbox{-}3B   & 80.84 & 0.56 & $-$0.382 & 84.26 & 0.44 & $-$0.382 & 54.93 & 0.94 & $-$0.418 \\
\rowcolor{blue!10}
\OURS\mbox{-}8B   & \textbf{85.18} & \textbf{0.52} & \textbf{$-$0.371} & \textbf{85.93} & \textbf{0.42} & \textbf{$-$0.374} & 56.21 & 0.93 & $-$0.406 \\
\bottomrule
\end{tabular}
}
\end{table*}

Antibody complementarity-determining region (CDR) design aims to predict the amino acid sequences of the CDR1, CDR2, and CDR3 regions given the framework region sequence of an antibody. CDR regions are the core functional regions responsible for specific antigen binding, and accurate prediction of their sequences is central to antibody engineering and therapeutic antibody design~\citep{schroeder2010structure}. Similar to protein editing, this task lacks a directly corresponding sequence format in the pre-training corpus, providing an additional test of downstream generalization. 
Notably, our evaluation is conducted under the inverse-folding setting, and most of the compared baselines are also inverse-folding methods. We adopt this challenging setup to probe the performance boundary of our sequence-based generative paradigm under a stringent comparison regime. Concretely, RADAb~\citep{wang2024retrieval}, for example, conditions generation on the 3D context of the antigen--antibody complex, the backbone geometry of the target CDR loop, and structurally retrieved CDR-like fragments from a protein database. These models therefore benefit from explicit geometric constraints that are unavailable to sequence.

\textbf{Setting.} We adopt the established antibody evaluation dataset SAbDab~\citep{dunbar2014sabdab} for assessment, fine-tuning and evaluating on its standard training/test split. Evaluation metrics include: AAR (Amino Acid Recovery, the per-position amino acid identity between the generated and ground-truth CDR sequences); scRMSD (Self-consistency RMSD, the C$\alpha$ RMSD in the CDR region between the structure refolded from the generated sequence via ABodyBuilder2~\citep{abanades2023immunebuilder} and the original antibody structure); and Plausibility (the pseudo-log-likelihood computed by the antibody language model AntiBERTy~\citep{ruffolo2021deciphering}). All metrics are reported separately for the heavy chain (CDR-H1/H2/H3) and light chain (CDR-L1/L2/L3). In terms of grammar design, the input uses \texttt{<CDR>} tokens to mark CDR positions within the antibody sequence, with \texttt{<CDRDesign>} as the task semantic token: \texttt{<AntibodyS>...<CDR>...<CDR>...<CDR>...<AntibodyE><CDRDesign>}. The output demarcates the three CDR sequences using chain-type-specific boundary tokens: the heavy chain takes the form \texttt{<AntibodyS>...<H1S>...<H1E>...<H2S>...<H2E>...<H3S>...<H3E>...<AntibodyE>}, and the light chain \texttt{<AntibodyS>...<L1S>...<L1E>...<L2S>...<L2E>...<L3S>...<L3E>...<AntibodyE>}. This design distinguishes heavy- and light-chain CDR regions through chain-type-specific boundary tokens while reusing the antibody modality grammar \texttt{<AntibodyS>...<AntibodyE>}.

\textbf{Baselines.} All baseline methods are inverse folding approaches that take the three-dimensional backbone structure of the antibody as input, including ProteinMPNN~\citep{dauparas2022robust}, ESM-IF1~\citep{hsu2022learning}, DiffAb-fix~\citep{luo2022antigen}, AbMPNN~\citep{dreyer2023inverse}, RADAb~\citep{wang2024retrieval}. The Grafting performance is directly derived from RADAb.

\begin{figure}[th]
    \centering
    \includegraphics[width=\textwidth]{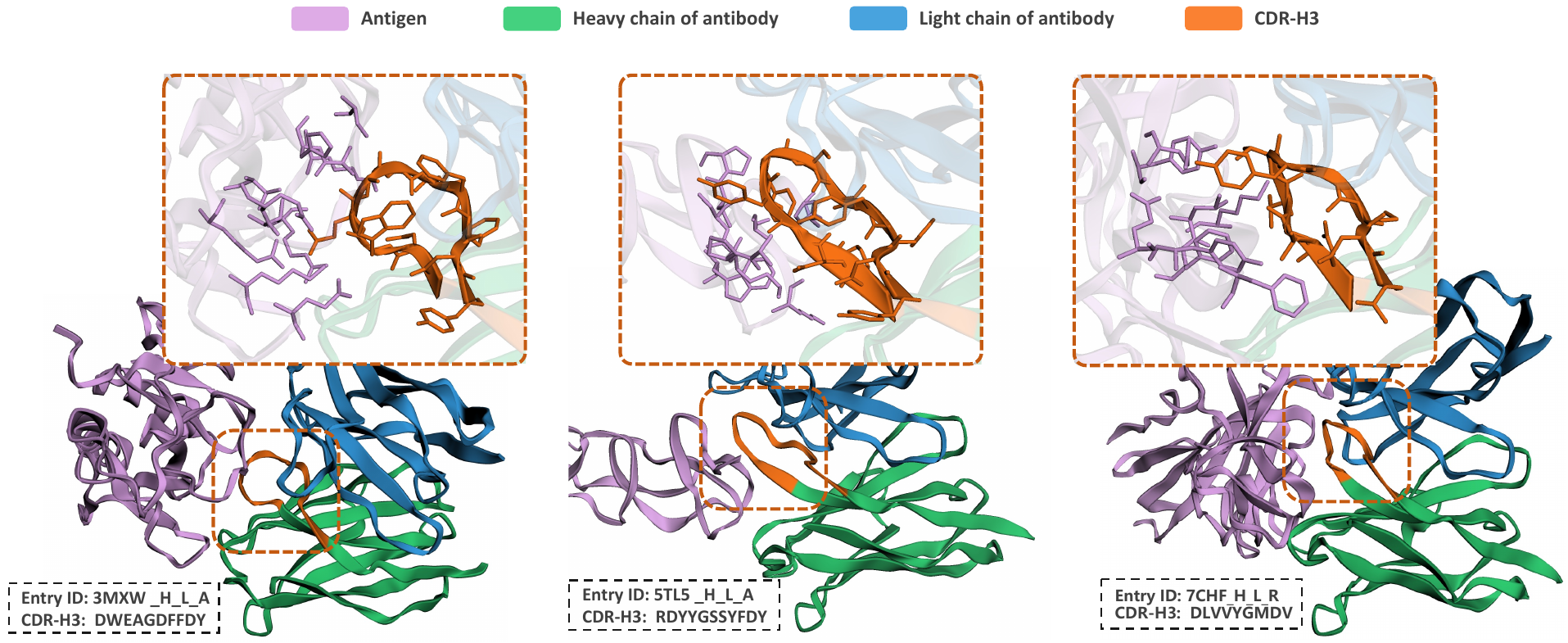}
    \caption{Visualization of antibody-antigen complexes from three SAbDab test cases, formed by LOGOS-generated antibodies and their corresponding antigens. Insets (dashed boxes) show the binding interface between the CDR-H3 region and the antigen.}
    \label{fig:antibody}
\end{figure}

\textbf{Results and Analysis.} As shown in Table~\ref{tab:sabdab}, LOGOS achieves the best performance on CDR1 and CDR2 regions across both AAR and scRMSD, exhibits a gap relative to inverse folding methods on CDR3, and attains competitive Plausibility across all regions.

\textit{Performance on CDR1 and CDR2 regions.} Across the four CDR1 and CDR2 loops (H1, H2, L1, and L2), LOGOS-8B achieves the highest AAR and the lowest scRMSD among the compared methods, with AAR exceeding 85\% on both CDR-L1 and CDR-L2. This trend is consistent with the relatively conserved nature of CDR1 and CDR2, whose sequences and conformations are largely influenced by germline V-gene segments and the surrounding framework context~\citep{north2011new,chothia1987canonical}. We speculate that pre-training on large-scale antibody sequence corpora helps LOGOS capture these conserved sequence–structure patterns, thereby improving CDR1/2 prediction from sequence alone.

\textit{Gap on CDR3 regions.} The CDR3 region—especially CDR-H3—is the region with the highest sequence diversity and greatest length variation in antibodies. For CDR-H3, its sequence is largely determined by the selection of D gene segments and random insertions/deletions at junctional regions during V(D)J recombination, rather than being predictable from framework-region sequence context alone~\citep{schroeder2010structure,xu2000diversity}. Consequently, sequence prediction for CDR3, particularly CDR-H3, is less tractable from sequence context alone and can benefit more from explicit constraints from the antibody three-dimensional structure (especially the CDR-H3 loop conformation), which is an inherent advantage of inverse folding methods. As a purely sequential model, LOGOS has predictable limitations in CDR3 recovery accuracy in the absence of 3D structural input.

\textit{Sequence plausibility.} LOGOS-8B achieves the best or highly competitive Plausibility scores across all six CDR regions. Since Plausibility is evaluated by the independent antibody language model AntiBERTy, this suggests that sequences generated by LOGOS generally conform to natural antibody distributions, even when their AAR on CDR3 is lower. This observation aligns with a known characteristic of CDR3 evaluation: given the high intrinsic sequence diversity of CDR, especially CDR-H3, multiple distinct sequences can constitute valid bindings for a given framework. AAR, which measures position-wise identity against a single reference, fails to fully capture this multiplicity. The competitive Plausibility of LOGOS on CDR3 regions supports the interpretation that the model generates distributionally plausible alternatives rather than merely recovering the specific reference sequence. Figure~\ref{fig:antibody} provides a structural visualization of three test cases, where the LOGOS-generated CDR-H3 loops form spatially coherent binding interfaces with their corresponding antigens.

\textit{Generalization beyond pre-training formats.} Together with protein editing, CDR design provides a second validation that sequence knowledge acquired under the unified scientific grammar can transfer to task formats absent from the pre-training corpus through minimal format adaptation.

\section{Conclusion and Future Work}

In this work, we propose LOGOS, a multi-domain generative framework that encodes heterogeneous scientific objects including proteins, antibodies, small molecules, chemical reactions, materials, and their spatial interactions into a shared discrete token space through a unified scientific grammar. By operating directly on domain-native representations rather than relying on natural language as an intermediary, LOGOS maintains formal and objective alignment between pre-training and downstream tasks. Across six representative tasks spanning different scientific domains and task types, LOGOS achieves highly competitive performance within a purely sequential paradigm, providing an initial validation of the "one model fits all" premise in the natural sciences.

In the future, several directions remain open for extending this framework.

\textbf{Domain coverage.} Limited by available computational resources, the current framework has not yet incorporated nucleic acid-related modalities such as genomic and transcriptomic sequences. Extending the scientific grammar to these domains is an important next step toward broader coverage of the natural sciences.

\textbf{Data and model scale.} The pre-training corpus covers a subset of publicly available data in each domain, and experiments span a parameter range of 1B to 8B where stable scaling behavior is observed. Scaling both data and model size would further test the capacity boundaries of the framework and help characterize scaling behavior on scientific modalities.

\textbf{Information modality.} The current design encodes spatial relationships through discretized token representations. Incorporating explicit geometric information to complement sequential modeling may further benefit tasks sensitive to three-dimensional structure.

Collectively, these directions point beyond the current feasibility study toward a broader long-term goal: a truly general-purpose scientific foundation model for unified understanding, prediction, and design across domains, scales, and modalities.

\clearpage
\bibliography{biblio}
\bibliographystyle{colm2024_conference}

\appendix
\section*{Appendix}
\addcontentsline{toc}{section}{Appendix}
\section{Overview of data construction}
\label{Overview of data construction}

\begin{figure}[th]
    \centering
    \includegraphics[width=\textwidth]{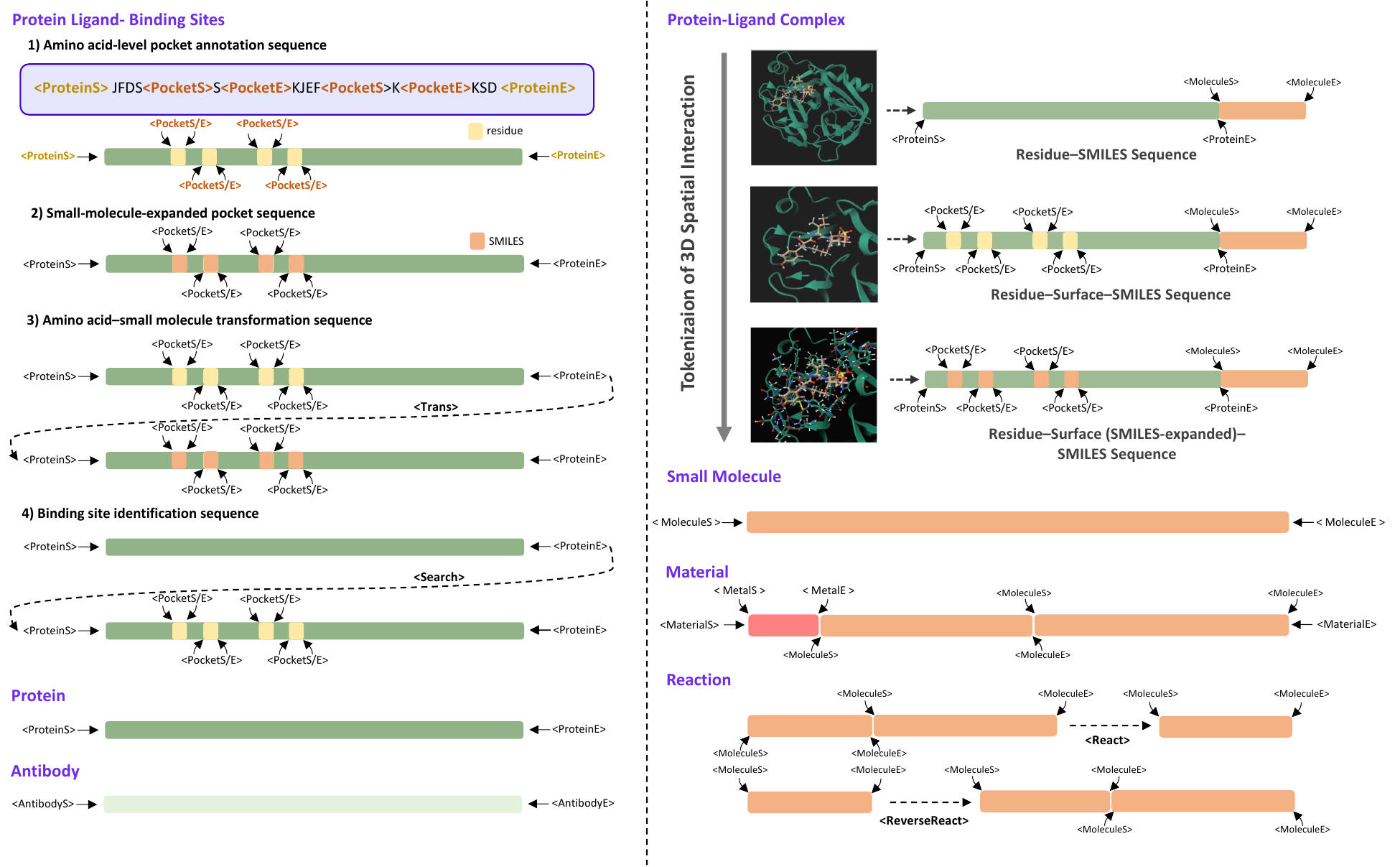}
    \caption{Overview of data construction in LOGOS. Colored bars distinguish the serialized representations of different scientific modalities—proteins (green), antibodies (light green), small molecules (orange), and metal clusters (red). Using protein ligand-binding sites and protein–ligand complexes as examples, it further illustrates how spatial interactions are discretized and incorporated into the unified scientific grammar, thereby encoding scientific objects from multiple domains and their relationships as sequences in a shared token space. }
    \label{fig:data construction}
\end{figure}

Figure~\ref{fig:data construction} gives a schematic summary of the data organization adopted in LOGOS. Under the unified scientific grammar, the model is designed to learn not from isolated single-modality data, but from scientific objects across multiple domains together with their relationships. In this framework, heterogeneous data sources with different representational forms are converted into discrete sequences in a shared token space, allowing diverse scientific modalities to be modeled within a common sequential representation. The colored bars in the figure correspond to the serialized forms of different modalities and provide an intuitive view of this unified organization.
A key aspect of this data system is the explicit incorporation of interaction-aware representations, especially for protein ligand-binding sites and protein–ligand complexes. For protein ligand-binding sites, multiple sequence forms are constructed to capture complementary aspects of the same object, including positional annotation within the protein sequence, residue expansion into side-chain chemical representations, and transformation between amino-acid-level and small-molecule-level forms. These designs establish an explicit connection between protein residue representations and small-molecule chemical representations. For protein–ligand complexes, protein pockets and ligands are organized into joint serialized representations that encode interface-level structural constraints and interaction patterns.
More importantly, this design enables spatial relationships that are usually described in terms of three-dimensional structures to be discretized and expressed as token sequences under the unified scientific grammar. As a result, cross-object interaction information can be incorporated into the same autoregressive sequence modeling framework as other scientific modalities. This data organization therefore serves as the foundation of LOGOS and supports cross-modal knowledge integration, relationship modeling, and unified generation across multiple scientific tasks.

\section{Evaluation metrics for different tasks}
In this section, we provide a detailed introduction to the evaluation metrics used for different tasks.

\subsection{Interaction-Aware Ligand Design for Binding Pockets}

In this task, we evaluate the quality of generated ligands from multiple aspects, including binding affinity, drug-likeness, synthetic accessibility, and physicochemical properties. Specifically, we report Vina score, QED, SAS, LogP, and LPSK.

\textbf{Vina Score.} The Vina score is derived from AutoDock Vina, a widely used molecular docking program~\citep{trott2010autodock}. It estimates the binding free energy (in kcal/mol) between a generated ligand and its target protein pocket. Lower values indicate stronger predicted binding affinity. In our evaluation, we report the average Vina score across multiple sampled ligands per method. Since lower scores correspond to better binding, this metric is marked with a downward arrow (↓), indicating that smaller values are preferable.

\textbf{Quantitative Estimate of Drug-likeness (QED).} QED is a composite metric ranging from 0 to 1 that quantifies how “drug-like” a molecule is, based on eight physicochemical properties including molecular weight, logP, hydrogen bond donors/acceptors, polar surface area, rotatable bonds, aromatic rings, and formal charge~\citep{bickerton2012quantifying}. Higher QED values (closer to 1) indicate greater similarity to known oral drugs. This metric helps filter out molecules that may bind well but are unlikely to succeed in later stages of drug development due to poor pharmacokinetic profiles. We mark this metric with an upward arrow (↑), as higher values are desired.

\textbf{Synthetic Accessibility Score (SAS).} SAS estimates how easy or difficult it is to synthesize a given molecule, based on fragment complexity and rarity observed in large chemical databases (e.g., ChEMBL~\citep{gaulton2012chembl})~\citep{ertl2009estimation}. Scores range from approximately 1 (very easy to synthesize) to 10 (very difficult). In practice, scores below 6 are generally considered synthetically feasible. For consistency with other metrics where higher is better, we invert the scale so that higher SAS values indicate easier synthesis — hence the upward arrow (↑).

\textbf{LogP (Partition Coefficient).} LogP measures the partition coefficient of a compound between octanol and water, reflecting its lipophilicity. Optimal LogP values for orally bioavailable drug candidates typically fall within the range of -0.4 to 5.6, as noted by ~\citet{lin2025cbgbench}. Values outside this interval may suggest poor absorption (too hydrophilic) or excessive accumulation in fatty tissues (too lipophilic). Unlike other metrics, LogP has no directional preference within the optimal window — both very low and very high values are undesirable. Thus, we do not assign an arrow direction to this metric. Instead, we report raw values and interpret them contextually against the recommended therapeutic range.

\textbf{LPSK.} The LPSK metric measures the proportion of generated molecules that satisfy Lipinski’s Rule of Five~\citep{lipinski2004lead}, which is a widely used guideline for assessing drug-like properties. This rule is based on several key physicochemical properties, including molecular weight, hydrogen bond donors, hydrogen bond acceptors, and LogP, and is commonly used to evaluate whether a molecule is likely to exhibit favorable characteristics as a drug candidate. Therefore, a higher LPSK value indicates that a larger proportion of the generated molecules satisfy these standard drug-likeness criteria.

\subsection{Protein Ligand-Binding Site Identification}

\textbf{Top-$n$ and Top-($n$+2).} Top-$n$ and Top-($n$+2) are ranking-based metrics for binding site identification. The evaluation follows the DCC criterion, where a predicted pocket is considered correct if the distance between the pocket center and any atom of a relevant ligand is within 4 Å~\citep{chen2011critical,krivak2018p2rank}. Since a protein structure may contain multiple relevant ligands, the number of considered predictions is adjusted accordingly. Specifically, for a structure with n relevant ligands, Top-n evaluates whether the relevant binding sites are successfully identified within the top n predicted pockets, while Top-(n+2) extends the range to the top n+2 predictions. Each relevant ligand contributes equally to the final success rate. These metrics therefore assess not only whether true binding sites are predicted, but also whether they are ranked near the top of the prediction list. Higher values indicate better pocket identification performance.

\subsection{Pocket Translation Task}
\label{appendix:translation}

\textbf{Site Acc.} Site accuracy adopts a single amino acid residue position within a pocket as the basic unit of evaluation. We pool together all amino acid residue positions from every sample in the test set, independently judge the correctness of each residue position according to the criterion described above, and report the proportion of residue positions that are correctly translated. 

\textbf{Sample Acc.} Sample accuracy adopts an entire pocket (i.e., a single evaluation instance together with all of its aligned residue positions) as the basic unit of evaluation. A pocket sample is judged as correct if and only if every one of its residue positions is correctly translated; as long as any single residue position is mismatched, missing, or unparseable, the sample as a whole is counted as incorrect. We then report the proportion of pocket samples in the test set that are fully and correctly translated.

\subsection{Retrosynthesis Prediction}

\textbf{Top-$k$.} Top-$k$ is a standard evaluation metric for retrosynthesis prediction. For each target product, the model generates a ranked list of candidate reactant sets, and a prediction is considered correct if the ground-truth reactant set appears within the top $k$ candidates. Therefore, top-$k$
measures the proportion of test examples for which the correct reactant set is successfully recovered among the top-$k$ predictions. This metric reflects both prediction accuracy and ranking quality. In this work, we report Top-1 and Top-3, corresponding to the cases where $k$ = 1 and $k$ = 3, respectively~\citep{xia2025nature}. Higher values indicate better retrosynthesis prediction performance.

\subsection{Unconditional Material Generation}

For MOF generation, we follow the evaluation protocol of MOFFlow-2~\citep{kim2026flexible} and report three metrics: Valid, VNU, and NBB. 

\textbf{Valid.} Valid measures the proportion of generated MOFs that pass the chemical validity check of MOFChecker~\citep{jin2025mofchecker}. 

\textbf{VNU.} VNU measures the proportion of generated MOFs that are simultaneously chemically valid, structurally novel, and structurally unique, where novelty is defined by whether the MOFid~\citep{bucior2019identification} is absent from the training set and uniqueness is evaluated after deduplication. 

\textbf{NBB.} NBB measures the proportion of valid generated MOFs that contain at least one novel building block not observed in the training set. 

Together, these three metrics evaluate generation performance from three increasingly strict perspectives: chemical validity, structural novelty, and component-level innovation. Higher values indicate better performance.

\subsection{Protein Editing}

\textbf{Fitness.} Fitness measures the normalized fitness of the generated sequences and reflects the extent to which the model can improve the target property. 

\textbf{Diversity.} Diversity is defined as the median edit distance over all pairwise comparisons within the generated sequence set, and is used to quantify the variation among generated candidates. 

\textbf{Novelty.} Novelty is defined as the median of the minimum edit distance from each generated sequence to all starting sequences, and measures how far the generated sequences deviate from the input sequences. 

Together, these metrics evaluate optimization quality from three complementary aspects: functional improvement, intra-set diversity, and sequence-level departure from the starting points~\citep{kirjner2024improving}.

\subsection{Antibody CDR Design}

We adopt the established antibody benchmark SAbDab~\citep{dunbar2014sabdab} for evaluation, using its standard training/test split for fine-tuning and testing. To assess the accuracy and rationality of the generated antibody sequences, following ~\citet{wang2024retrieval}, we consider three widely used evaluation metrics.

\textbf{Amino Acid Recovery (AAR, \%).} AAR measures the sequence-level similarity between the designed CDR sequence and the ground-truth CDR sequence. It is defined as the proportion of positions at which the generated amino acid matches the native amino acid. Higher AAR indicates better sequence recovery.

\textbf{Self-consistency RMSD (scRMSD, \AA).} To evaluate structural consistency, we refold the generated antibody sequence using ABodyBuilder2~\citep{abanades2023immunebuilder}. We then align the refolded antibody framework to the original antibody structure and compute the RMSD of the C$\alpha$ atoms in the CDR region. Lower scRMSD indicates that the generated sequence is more compatible with the target antibody structure.

\textbf{Plausibility.} We evaluate the naturalness of the generated antibody sequence using AntiBERTy~\citep{ruffolo2021deciphering}. Specifically, plausibility is measured by the pseudo-log-likelihood assigned by AntiBERTy to the generated sequence. Higher plausibility indicates that the generated sequence is more consistent with the distribution of natural antibodies.

All metrics are reported separately for the heavy-chain CDRs (CDR-H1/H2/H3) and the light-chain CDRs (CDR-L1/L2/L3).

\section{Detailed configurations}

We conduct unified continual pre-training and post-training on multi-domain scientific data across models with different parameter scales and different backbones, with detailed configurations provided below.

\begin{table}[th]
\centering
\caption{Hyperparameters for continual pre-training of LOGOS. All training is conducted on 32 NVIDIA A800 GPUs.}
\label{tab:pretrain-hparams}
\begin{tabular}{ll}
\toprule
\textbf{Category} & \textbf{Value} \\
\midrule
Optimizer                      & AdamW \\
Peak learning rate             & $2.0\times10^{-5}$ \\
LR scheduler                   & Cosine \\
Warmup ratio                   & 0.06 \\
Max training steps             & 100000 \\
Precision                      & bfloat16\\
Batch size                     & 1024 \\
\bottomrule
\end{tabular}
\end{table}

\begin{table}[th]
\centering
\caption{Hyperparameters for post training of LOGOS. All training is conducted on 8 NVIDIA A800 GPUs.}
\label{tab:sft-hparams}
\begin{tabular}{ll}
\toprule
\textbf{Category} & \textbf{Value} \\
\midrule
Optimizer                      & AdamW \\
Peak learning rate             & $1.0\times10^{-5}$ \\
LR scheduler                   & Cosine \\
Warmup ratio                   & 0.1 \\
Precision                      & bfloat16 \\
Training epochs                & 8 \\
Batch size                     & 128 \\
\bottomrule
\end{tabular}
\end{table}

The optimal generation parameters for different downstream tasks during the inference phase are as follows. The values of top\_p, temperature, and repetition\_penalty are 0.85, 1.2, and 1.05 for the interaction-aware ligand design for binding pockets, retrosynthesis prediction, and unconditional material generation tasks, and 0.9, 1.2, and 1.0 for the protein ligand-binding site identification, protein editing, and antibody CDR design tasks.

\end{document}